\documentclass[11pt]{article}

\usepackage[a4paper,margin=1in]{geometry}
\usepackage{times}
\usepackage{graphicx}
\usepackage{amsmath, amsthm, amsfonts, amssymb, amscd, bm, cases,dsfont}
\usepackage{multirow}
\usepackage{booktabs}
\usepackage{hyperref}
\usepackage{authblk}
\usepackage{setspace}
\usepackage{caption}
\usepackage{url}
\usepackage{enumitem}
\usepackage{xcolor}
\usepackage{natbib}
\usepackage{xspace}
\usepackage{algorithm}
\usepackage{algorithmic}
\usepackage{subcaption}
\usepackage{wrapfig}

\newtheorem{lemma}{Lemma}

\newtheorem{definition}{Definition}

\newtheorem{remark}{Remark}

\title{\textbf{Constrained Adversarial Perturbation}}

\author[1]{Virendra Nishad}
\author[2]{Bhaskar Mukhoty}
\author[3]{Hilal AlQuabeh}
\author[4]{Sandeep K. Shukla}
\author[1]{Sayak Ray Chowdhury}
\affil[1]{IIT Kanpur, India}
\affil[2]{IIT Delhi, India}
\affil[3]{MBZUAI, UAE}
\affil[4]{IIIT Hyderabad, India}
\date{}

\hypersetup{
    colorlinks=true,
    linkcolor=blue,
    citecolor=blue,
    urlcolor=teal,
}

\newcommand{\capx}{CAPx\xspace}
\newcommand{\capX}{\textsc{CAPX}\xspace}

\newcommand{\norm}[1]{\lVert{#1}\rVert}
\newcommand{\R}{\mathbb{R}}
\newcommand{\Rplus}{\mathbb{R}_{\geq 0}}

\newcommand{\X}{\mathcal{X}}

\newcommand{\ASR}{\mathrm{ASR}}
\newcommand{\sign}{\mathrm{sign}}

\DeclareMathOperator*{\argmax}{arg\,max}

\begin{document}
\maketitle

\begin{abstract}
Deep neural networks have achieved remarkable success in a wide range of classification tasks. However, they remain highly susceptible to adversarial examples—inputs that are subtly perturbed to induce misclassification while appearing unchanged to humans. Among various attack strategies, Universal Adversarial Perturbations (UAPs) have emerged as a powerful tool for both stress-testing model robustness and facilitating scalable adversarial training. Despite their effectiveness, most existing UAP methods neglect domain-specific constraints that govern feature relationships. Violating such constraints—such as debt-to-income ratios in credit scoring or packet-flow invariants in network communication—can render adversarial examples implausible or easily detectable, thereby limiting their real-world applicability.

In this work, we advance universal adversarial attacks to constrained feature spaces by formulating an augmented Lagrangian–based min-max optimization problem that enforces multiple, potentially complex constraints of varying importance. We propose Constrained Adversarial Perturbation (CAP), an efficient algorithm that solves this problem using a gradient–based alternating optimization strategy. We evaluate CAP across diverse domains—including finance, IT networks, and cyber-physical systems—and demonstrate that it achieves higher attack success rates while significantly reducing runtime compared to existing baselines. Our approach also generalizes seamlessly to individual adversarial perturbations, where we observe similar strong performance gains. Finally, we introduce a principled procedure for learning feature constraints directly from data, enabling broad applicability across domains with structured input spaces.
\end{abstract}

\section{Introduction}
Deep learning based classification algorithms have achieved remarkable success across a diverse range of applications, including image classification, loan default prediction, malicious activity detection in IT networks, and autonomous decision-making. Despite these advancements, deep classifiers remain highly susceptible to adversarial examples (AEs)—inputs that have been subtly and deliberately perturbed to cause incorrect model predictions \cite{szegedy2013intriguing}. A growing body of work highlights the alarming vulnerability of these classifiers to adversarial examples in high-stakes, real-world scenarios. For example, strategically placed stickers on a physical stop sign can mislead computer vision models into classifying it as a speed limit sign, despite its appearance remaining unambiguous to human observers \cite{eykholt2018robust}; Specially crafted eyeglass frames can enable an attacker to impersonate any individual in a facial recognition system compromising biometric security \cite{sharif2016accessorize}; imperceptible perturbations to radiographic images can lead diagnostic models to produce dangerously incorrect medical predictions \cite{finlayson2019adversarial}.

Under the white-box threat model---assumes full access to the target architecture and parameters, adversarial perturbations are generated either individually or universally. The individual perturbation approach crafts input-specific modifications designed to cause a particular example to be misclassified \cite{szegedy2013intriguing}. In contrast, the universal perturbation approach utilizes a set of training examples to compute a single perturbation vector, which, when added to unseen test inputs, induces misclassification with high confidence \cite{moosavi2017universal}. The primary distinction between the two methods lies in their generalizability: individual perturbations require knowledge of the exact input to be perturbed, whereas universal perturbations are input-agnostic and work without access to the test instance.

Universal adversarial perturbation (UAP) has emerged as a versatile approach for evaluating and enhancing the robustness of neural network classifiers. For instance, UAP has been proposed as a scalable defense mechanism capable of operating on large-scale datasets such as ImageNet \cite{shafahi2020universal, pan2024adversarial} as well as on a wide range of transformations in the physical world \citet{xu2024robust}. In black-box settings—where access to model weights is limited—studies have demonstrated that UAPs can still achieve high attack success rates, even under strict query limitations \cite{zhao2019universal}. 
Moreover, research has extended beyond purely additive perturbations to explore spatial, generative, and class-discriminative variants of UAPs, offering deeper insights into model interpretability and sensitivity \cite{papernot2016limitations, zhang2020understanding}.

The effectiveness of many of these prior adversarial methods is largely confined to domains (e.g., images) where there are no explicit constraints on the features. However, constraints over the feature space play a critical role in many application domains, e.g., debt-to-income ratios in credit scoring, packet-flow invariants in network communication, etc. Of particular importance are safety-critical systems, where the feature space is defined by measurements from sensors and actuators collected during normal system operation, and the measurements are used to train classifiers for detecting anomalies. Importantly, these systems leverage physical invariants—domain-specific relationships among sensor and actuator values that consistently hold under nominal conditions—as an independent layer of anomaly detection \cite{quinonez2020savior}. By combining machine learning with invariant-based validation, these systems enhance overall reliability and robustness \cite{pang2021deep}.
However, adversarial methods that disregard such constraints often perform poorly in these domains: while they may succeed in inducing misclassification, it comes at the cost of violating fundamental feature dependencies, rendering the perturbed inputs implausible or easily detectable. Against this backdrop, we advance the study of adversarial perturbations to \emph{constrained feature spaces}, incorporating domain-specific structure directly into the adversarial optimization framework.

\paragraph{Contributions}
\begin{itemize}
    \item  We formulate adversarial perturbation under feature constraints as a min-max optimization problem that jointly captures the objectives of inducing misclassification and adhering to domain-specific constraints. The min-max formulation enables computing minimum $\ell_2$-norm perturbations, while explicitly penalizing constraint violations via augmented Lagrange multipliers.
    
    \item The explicit penalty variables offer fine-grained control over constraint enforcement and allow incorporation of domain knowledge (e.g., relative importance or strictness of different constraints) into the optimization process. This flexible and principled approach supports optimization under complex, high-dimensional constraints often encountered in real-world systems.

    \item We propose CAPX, a novel universal adversarial perturbation algorithm under constrained feature spaces, that solves the proposed min-max optimization problem with a gradient–based alternating optimization strategy. The algorithm is parallelizable on modern GPU architectures, resulting in substantial reductions in runtime and improved scalability without compromising accuracy.

    \item We evaluate CAPX on datasets spanning a diverse range of domains, including loan processing, IT network traffic analysis, medical diagnostics, and cyber-physical systems. Empirical results demonstrate that CAPX consistently outperforms baselines, achieving absolute gains varying from 2.90\% to 47.90\% in attack success rate relative to the strongest baseline. Additionally, CAPX offers substantial computational advantages, exhibiting over a minimum of 45 times reduction in runtime compared to the most efficient existing approach across all datasets.
    
    \item Our framework naturally extends to individual adversarial perturbations, for which we introduce CAPx---a variant tailored to input-specific attacks. CAPx also enjoys strong performance gains across all evaluated domains. In addition, we propose a principled procedure for learning feature constraints directly from data, enabling the broad applicability of our approach across domains where constraints are not explicitly specified. 
\end{itemize}

\paragraph{Notations} Throughout the paper, lower-case letters represent scalars, bold lower-case letters represent vectors, and upper-case letters represent matrices and sets. The symbol $\partial$ denotes the partial derivative, $\nabla$ denotes the gradient, and $[n]$ represents a sequence of numbers from 1 to $n$.

\section{Background and Problem Setup} \label{sec:background_and_problem_setup}

In this section, we formalize the notion of constrained adversarial perturbations. Let $(\bm{x},y) \in \mathcal{X} \times \mathcal{Y}$ be an example, where $\mathcal{X}\subseteq \mathbb{R}^d$ is a $d$-dimensional feature space and $\mathcal{Y} = \{ +1, -1\}$ is a binary label space. Let $f: \R^d \to \R$ be an arbitrary scalar-valued classification function. For example, if $f$ is an affine classifier, then $f(\bm{x})= \bm{w}^\top \bm{x} +b$ for some $\bm{w} \in \R^d$ and $b \in \R$. In general, $f$ can be any differentiable function (e.g., neural nets). 

A hypothesis of the form $ \bm{x} \mapsto \sign(f(\bm{x}))$ labels positively all points falling on one side of the level set $f(\bm{x})=0$ and negatively all others. Thus $\bm{x}$ is classified correctly when $y\cdot f(\bm{x}) > 0$ and incorrectly otherwise. Given the feature vector $\bm{x}$, an (unconstrained) adversarial attack aims to find a minimum-norm perturbation $\bm{\delta} \in \R^d$
that fools the classifier to label $\bm{x}$ incorrectly by inducing $y\cdot f(\bm{x}+\bm{\delta}) < 0$, while ensuring $\bm{x}+\bm{\delta} \in \X$.
\begin{remark}
Our formulation can be readily extended to the multiclass classification problem 
using a one-vs-all approach. In this setting, the classifier $\bm{f}:\R^d \to \R^m$ is a vector-valued function, where $m$ is the number of classes and the hypothesis is of the form $\bm{x} \mapsto \argmax_k \bm{f}_k(\bm{x})$.
\end{remark}
In this work, we investigate adversarial perturbations in structured feature spaces, where domain-specific dependencies and relationships exist among features. We formalize this notion by introducing generic constraints over the feature space, as defined below.

\begin{definition}[Feature constraints] \label{def:feat_cons}
Let $\bm{g}: \mathbb{R}^d \rightarrow \mathbb{R}^q$ be a vector-valued differentiable function and $\bm{b} \in \mathbb{R}^q$ be a vector. Then, for a feature vector $\bm{x}$, the constraints are given by
\begin{equation} \label{eq:def_feat_cons}
    \bm{g}(\bm{x}) \leq \bm{b}~,   
\end{equation} 
where the inequality holds element-wise and each $g_j(\bm{x}) \le b_j$, $j \in [q]$, represents a single constraint on the feature $\bm{x}$.  
\end{definition}
We place no restrictions on the function $\bm{g}$ beyond differentiability, making Definition~\ref{def:feat_cons} highly general. As a result, it captures linear constraints like feature boundaries and dependencies (e.g., $x_i > 0$, $x_i+x_j \le 0$, etc.), as well as non-linear constraints like numerical relationships between features (e.g., $x_i/x_j \le x_k$, $\max \{ x_i,x_j\}$, etc.) – forms of constraints that have been studied in the literature~\cite{papernot2016limitations,li2021conaml, simonetto2021unified}.

\paragraph{Threat Model}
We consider a white-box threat model, where the adversary has full access to the classifier $f$ (e.g., its architecture, parameters, loss function), examples $(\bm{x},y)$, and feature constraints $\bm{g}, b$. We assume that the adversary can perform unrestricted, gradient-based optimization via backpropagation. This setting simulates the worst-case scenario, with complete visibility into the model's internal components. It can also directly perturb the features, but needs to ensure the perturbed features satisfy the constraints~\eqref{eq:def_feat_cons}.

\paragraph{Constrained Adversarial Perturbation} For an example $(\bm{x},y)$, classifier $f$, and constraints $\bm{g},\bm{b}$, we define an adversarial attack as the minimum $\ell_2$-norm perturbation $\bm \delta$ that satisfies feature constraints while forcing misclassification:
\begin{align}\label{eq:cap}
        \underset{\bm{\delta} \in \R^d}{\min} \,  \frac{1}{2}\|\bm{\delta}\|_2^2\,\,
        \text{s.t.} \,\, \bm{g}(\bm{x} + \bm{\delta}) \leq \bm{b},\,\, 
         y\cdot f(\bm{x} + \bm{\delta}) < 0~.
\end{align}
The primary objective of this work, informally, is to compute a universal perturbation that solves the optimization problem in \eqref{eq:cap} simultaneously across all examples $(\bm{x}, y)$ drawn from an unknown data distribution.\footnote{Formally, we want this to hold with high probability over the data distribution. The formulation also generalizes to $\ell_p$-norms for any $p \ge 1$.} The adversary is provided with a training set $\bm{X} = \{\bm{x}_i\}_{i=1}^n$ consisting of $n$ examples (features) from a common class $y \in \mathcal{Y}$ and aims to construct a perturbation $\bm{\delta}_{\bm{X}}$ that maximizes the attack success rate (ASR) on the set $\bm{X}$, as formally defined below.
\begin{definition}[ASR]
The attack success rate of a perturbation $\bm{\delta}$ over a set $\bm{X} = \{\bm{x}_i\}_{i=1}^{n}$ from class $y$ is given by
\begin{equation*}
    \ASR(\bm{\delta}) = \frac{1}{n}\sum_{i=1}^n \mathds{1} \lbrace ( \bm{g}(\bm{x}_i + \bm{\delta}) \leq \bm{b} ) \cap (y\cdot f(\bm{x}_i + \bm{\delta}) < 0)\rbrace~,
\end{equation*}    
where $\mathds{1}\{\cdot\}$ is the indicator function.
\end{definition}
The underlying assumption behind the universal perturbation objective is standard: a perturbation $\bm{\delta}_{\bm{X}}$ achieving high ASR on the training set $\bm{X}$ is expected to generalize to unseen test examples drawn from the same distribution. In contrast, the individual perturbation objective aims to compute a distinct perturbation $\bm{\delta}_{\bm{x}}$ for each test example $(\bm{x}, y)$ such that the optimization problem in \eqref{eq:cap} is satisfied with $\ASR = 1$ for that specific instance.

\subsection{Related Work}

Existing adversarial attack methods often solve the individual perturbation problem by maximizing the loss function of the classifier over the perturbation $\bm{\delta}$, subject to the constraint that $\bm{\delta}$ lies within an $\ell_p$-norm ball of radius $\epsilon$. Notable approaches include PGD \cite{madry2018towards}, which performs iterative gradient ascent with projection onto the $\ell_p$-ball, and the Fast Gradient Sign Method \cite{goodfellow2014explaining} that considers the first-order approximation of the loss function around the current perturbed input with an $\ell_\infty$-ball projection.
DeepFool \cite{moosavi2016deepfool} 
finds the minimum $\ell_p$-norm perturbation that fools the classifier to change its prediction, i.e., a $\bm{\delta}_{\bm{x}}$ that enforces $\sign(f(\bm{x}+\bm{\delta}_{\bm{x}})) \neq \sign(f(\bm{x}))$.
These methods require access to the specific input $\bm{x}$ and assume no structural constraints among features, limiting their applicability in domains with explicit feature dependencies.

The Universal Adversarial Perturbation (UAP) method \cite{moosavi2017universal} learns a single perturbation $\bm{\delta}_{\bm{X}}$ over a training set $\bm{X}$ such that $\sign(f(\bm{x}_i + \bm{\delta}_{\bm{X}})) \neq \sign(f(\bm{x}_i))$ for most $\bm{x}_i \in \bm{X}$. UAP updates $\bm{\delta}$ by applying DeepFool to misclassify samples but does not enforce any feature constraints, limiting its applicability in structured domains. ConAML \cite{li2021conaml} extends UAP to constrained settings by updating $\bm{\delta}$ via gradient ascent while enforcing linear constraints through a partitioning of features into independent and dependent sets. However, it remains restricted to linear constraints, and no prior method, to our knowledge, addresses universal perturbation under general nonlinear constraints. Moreover, both UAP and ConAML exhibit oscillatory behavior during optimization as gradients computed across different samples often point in conflicting directions. This leads to unstable updates and often results in suboptimal convergence.

MoEvA2 \cite{simonetto2021unified} is a grey-box, multi-objective evolutionary attack that supports non-linear constraints by jointly optimizing misclassification, $\ell_2$-norm perturbation, and constraint satisfaction using the R-NSGA-III genetic algorithm \cite{vesikar2018reference}. While originally designed for individual perturbations $\bm{\delta}_{\bm{x}}$, MoEvA2 can, in principle, be integrated into the UAP framework by replacing DeepFool. However, its high computational complexity, combined with the iterative nature of UAP, results in a computationally prohibitive procedure for generating universal perturbations under constraints. These limitations underscore the need for a method that can generate universal perturbations under generic constraints of the form $\bm{g}(\bm{x}) \le \bm{b}$, has low computational complexity, and scales efficiently to large datasets.

\section{Proposed Method} \label{sec:proposed_method}
In this section, we propose a method to solve the Constrained Adversarial Perturbation (CAP) problem~\eqref{eq:cap} over a training set $\bm{X}$ and refer to it as \capX. 

We propose to obtain a universal adversarial perturbation $\bm{\delta}_{\bm{X}}$ by solving the following optimization problem:
\begin{subequations} \label{eq:X_obj}
    \begin{align}
    \underset{\bm{\delta} \in \R^d}{\min} \quad & \frac{1}{2}\|\bm{\delta}\|^2\\
        \text{s.t.} \ \forall \bm{x}_i \in \bm{X}, \quad & \bm{g}(\bm{x}_i + \bm{\delta}) \leq \bm{b}~, \\
        \,& y\cdot f(\bm{x}_i + \bm{\delta}) \leq c~.\label{eq:X_mclf_con}
    \end{align}
\end{subequations}

\noindent
Here $c$ is a small negative threshold (e.g., $c \approx -10^{-4}$) introduced to relax the strict inequality in \eqref{eq:cap}, thereby transforming the misclassification condition into a more tractable non-strict inequality. It's an input to the algorithm.

Equality constraints are generally easier to manage within an optimization framework. To this end, we introduce slack variables $\bm{\Phi} = \{\phi_{i,j}\}_{i,j}$, $\bm{\theta} = \{\theta_i\}_{i}$, $i\in [n], j \in [q]$, to reformulate feature constraints and misclassification condition \eqref{eq:X_mclf_con} as equalities, where each $\bm{\phi}_i \in \mathbb{R}^q$ and $\theta_i \in \mathbb{R}$. The resulting adversarial objective is given by:
\begin{subequations} \label{eq:X_obj_eq_con}
    \begin{align}
        \min_{\bm \delta \in \R^d} \quad & \frac{1}{2}\|\bm{\delta}\|^2 \label{eq:X_eq_obj}\\
        \text{s.t.} \ \forall \bm{x}_i \in \bm{X}, \quad & \bm{g}(\bm{x}_i + \bm{\delta}) + \bm{\phi}_i - \bm{b} = \bm 0 ~,\label{eq:X_eq_feat_con}\\
        & y\cdot f(\bm{x}_i + \bm{\delta}) + \theta_i - c = 0 ~,\label{eq:X_eq_mclf_con}\\
        & \phi_{i,j} \geq 0~, \theta_i \geq 0~,\forall i \in [n]~, j \in [q] ~.\label{eq:X_eq_slack_con}
    \end{align}
\end{subequations}
\noindent
Although the constraints in \eqref{eq:X_eq_slack_con}—introduced via slack variables—are inequalities, they are simple non-negativity conditions. For now, we omit them from the main formulation and address them separately later.

With only the equality constraints, \eqref{eq:X_eq_feat_con} and \eqref{eq:X_eq_mclf_con}, we can convert \eqref{eq:X_eq_obj} to the following unconstrained objective:
\begin{align}
    \mathcal{L} & (\bm{\delta}, \bm{\Lambda}, \bm{\mu}, \bm{\Phi}, \bm{\theta}, \bm{P}, \bm{p})  =  \frac{1}{2}\|\bm{\delta}\|^2 \nonumber\\
    & + \sum\nolimits_{i=1}^{n} \sum\nolimits_{j=1}^q \lambda_{i,j} \cdot \lbrace g_j(\bm{x}_i + \bm{\delta}) + \phi_{i,j}  - b_j \rbrace  
    + \sum\nolimits_{i=1}^{n} {\mu_i} \cdot \lbrace y \cdot f(\bm{x}_i + \bm{\delta}) + \theta_i - c \rbrace   \\
    & + \frac{1}{2}\sum\nolimits_{i=1}^{n} \sum\nolimits_{j=1}^{q} \rho_{i,j} \cdot \lbrace g_j(\bm{x}_i + \bm{\delta}) + \phi_{i,j}  - b_j \rbrace^{2} 
    + \frac{1}{2}\sum\nolimits_{i=1}^{n}\varrho_i \cdot \lbrace y\cdot f(\bm{x}_i+\bm{\delta}) + \theta_i - c \rbrace ^2~. \nonumber\label{eq:X_ALF}
\end{align}
Here $\bm{\Lambda} = \lbrace\lambda_{i,j}\rbrace_{i,j}$, $\bm{\mu} = \lbrace\mu_i\rbrace_{i}$ are Lagrange multipliers associated with feature and misclassification constraints respectively, with $\lambda_{i,j}, \mu_i \in \mathbb{R}$, and $\bm{P} = \{\rho_{i,j}\}_{i,j}$, $\bm{p} = \{\varrho_i\}_{i}$ are the penalty variables associated with feature and misclassification constraints respectively, with $\rho_{i,j}, \varrho_i  \in \mathbb{R}_{+}$. 

The objective in \eqref{eq:X_ALF} is known as the augmented Lagrangian function (ALF) \cite[Chapter 17]{nocedal2006numerical}.\footnote{In contrast to the standard ALF method, we allow different penalty parameters for each constraint to accommodate domain-specific requirements.} This formulation differs from the standard Lagrangian method by incorporating squared penalty terms, and from the quadratic penalty method by retaining the linear terms involving Lagrange multipliers. As a result, it inherits desirable properties from both approaches.

On one hand, the Lagrange multipliers $\lambda_{i,j}, \mu_i$ aid in satisfying the feasibility conditions \eqref{eq:X_eq_feat_con} and \eqref{eq:X_eq_mclf_con} without requiring the penalty parameters $\rho_{i,j}, \varrho_i \to \infty$. This allows feasibility to be achieved with moderate values of $\rho_{i,j}, \varrho_i$, preventing the objective in \eqref{eq:X_ALF} from becoming unbounded. On the other hand, the quadratic penalty terms guide the optimization trajectory toward the feasible manifold even when the initial estimates of the Lagrange multipliers are poor. Moreover, they contribute rank-one regularization terms of the form $\rho_{i,j}\nabla g_j(\bm{x}_i+\bm{\delta})\nabla g_j(\bm{x}_i+\bm{\delta})^\top$ and $\varrho_i \nabla f(\bm{x}_i+\bm{\delta})\nabla f(\bm{x}_i+\bm{\delta})^\top$ to the Hessian $\nabla^2_{\bm{\delta}} \mathcal{L}(\cdot)$, improving its conditioning. This regularization ensures better local convergence properties, particularly in non-convex settings where convexity of $f$ or $g_j$ with respect to $\bm{\delta}$ is not assumed.

Now, we tackle the constraints involving slack variables in \eqref{eq:X_eq_slack_con}. Taking partial derivative of \eqref{eq:X_ALF} with respect to $\phi_{i, j}$ and $\theta_i$ gives a necessary condition for optimality:
\begin{subequations}
    \begin{align}
        \lambda_{i,j} + \rho_{i,j} \cdot \lbrace g_j (\bm{x}_i + \bm{\delta}) + \phi_{i, j} - b_j \rbrace = 0~, \\
        \mu_i + \varrho_i \cdot \lbrace y\cdot f(\bm{x}_{i} + \bm{\delta}) + \theta_i - c\rbrace = 0~.
    \end{align}
\end{subequations}

\noindent Solving for the slack variables $\phi_{i, j}$, $\theta_i$ and enforcing non-negativity constraints ignored earlier gives
\begin{subequations}
    \begin{align}
        \label{eq:X_optim_slack}
        \phi_{i, j} =& \ \max \{0, \ - \lambda_{i, j} / \rho_{i, j} - g_j(\bm{x}_i + \bm{\delta}) + b_j \}~, \\
        \theta_i =& \ \max \{ 0, \ - \mu_i/\varrho_i - y \cdot f(\bm{x}_i + \bm{\delta}) + c \}~. 
    \end{align}
\end{subequations}
\noindent For notational convenience, we define
    \begin{align} \label{eq:def_of_zeta_n_xi}
        \zeta(\bm{\delta}, \lambda_{i, j}, \rho_{i, j}) &:= \max \{ g_j(\bm{x}_i + \bm{\delta}) - b_j, -\lambda_{i, j}/\rho_{i, j} \}~, \nonumber \\
        \xi(\bm{\delta}, \mu_i, \varrho_i) &:= \max \{ y\cdot f(\bm{x}_i + \bm{\delta})-c, -\mu_i/\varrho_i \}~.
    \end{align}
Then, the ALF objective in \eqref{eq:X_ALF} can be simplified as
\begin{align} \label{eq:X_simplified_ALF}
    \mathcal{L}(\bm{\delta}, \bm{\Lambda}, \bm{\mu}, \bm{P}, \bm{p}) & = \frac{1}{2}\|\bm{\delta}\|^2 
    + \!\sum_{i=1}^{n}\sum_{j=1}^q  \lambda_{i,j} \cdot \zeta(\bm{\delta}, \lambda_{i,j}, \rho_{i,j}) + \!\sum_{i=1}^{n} \mu_i \cdot \xi(\bm{\delta}, \mu_i, \varrho_i) \\
    & + \!\frac{1}{2} \!\sum_{i=1}^n\sum_{j=1}^q \rho_{i, j} \cdot \zeta(\bm{\delta}, \lambda_{i,j}, \rho_{i, j})^2 \!+ \!\frac{1}{2} \sum_{i=1}^n \varrho_i \cdot \xi(\bm{\delta}, \mu_i, \varrho_i)^{2}.\nonumber
\end{align}
With this, we propose to obtain the universal perturbation by solving the following primal-dual optimization problem 
\begin{equation}
    \min_{\bm{\delta}} \ \max_{\bm{\Lambda}, \bm{\mu}} \  \mathcal{L}(\bm{\delta}, \ \bm{\Lambda}, \ \bm{\mu}, \ \bm{P}, \ \bm{p})~.\label{eq:X_final_uncon_UAP_obj}
\end{equation}
The primal gradients $\nabla_{\bm{\delta}}\mathcal{L}(\cdot)$ admit a closed-form expression for linear classifiers, while for neural networks, it can be computed using automatic differentiation. The dual gradients 
admit the following closed-form expressions.

\begin{lemma}[Dual gradient] \label{le:dual_gradient}
The gradients of the ALF \eqref{eq:X_simplified_ALF} with respect to the dual variables are given by
    \begin{subequations} \label{eq:dual_gradients}
    \begin{align}
        \partial_{\lambda_{i,j}} \mathcal{L}(\bm{\delta}, \bm{\Lambda}, \bm{\mu}, \bm{P}, \bm{p}) &= \zeta(\bm{\delta}, \lambda_{i,j}, \rho_{i,j})~,\\
        \partial_{\mu_i} \mathcal{L}(\bm{\delta}, \bm{\Lambda}, \bm{\mu}, \bm{P}, \bm{p}) &= \xi(\bm{\delta}, \mu_i, \varrho_i)~.
    \end{align}
\end{subequations} 
\end{lemma}
\noindent We provide the proof for the Lemma \ref{le:dual_gradient} in the appendix.

\subsection{Algorithm: CAPX}

 \begin{algorithm}[tb]
\caption{\capX}\label{alg:CAPX}
\noindent\textbf{Input:} Training set $\{\bm{x}_i\}_{i=1}^n$, label $y$, classifier $f$, classification threshold $c \!<\! 0$,
constraints $\lbrace g_j,b_j \rbrace_{j=1}^q$, maximum iteration $K$, learning rate $\alpha \!>\! 0$, penalty scaling $\tau \!>\! 1$, penalty frequency update factor $r \!\le\! K$, penalty upper bound $\Bar{\upsilon} \!>\! 0$.

\noindent\textbf{Variable Initialization:}
    Primal: $\bm{\delta}^0 \gets \mathbf{0}^{d \times 1}$,
    Dual: $\lambda_{i,j}^0 \sim \mathcal{U}[0, 10^{-4}],\mu_i^0\sim \mathcal{U}[0, 10^{-4}]$, $i \!\in\! [n], j \!\in\! [q]$,\\
    Penalty: $\rho_{i,j}^0 \gets 1 ,\varrho_i^0  \gets 1$, $i \!\in\! [n], j \!\in\! [q]$.
    
\noindent\textbf{Learning Steps:}
\begin{algorithmic}[1]
    \WHILE{$k \le K$}

            \STATE $\!\bm{\delta}^{k} \!\gets\! \bm{\delta}^{k-1} \!-\alpha \nabla_{\bm{\delta}} \mathcal{L}(\bm{\delta}^{k-1}\!,\!\bm{\Lambda}^{k-1}\!, \bm{\mu}^{k-1}\!,\!\bm{P}^{k-1}\!, \bm{p}^{k-1})$ \hfill \textit{// primal variable $\bm{\delta}$ update} 
         
        \FOR{$i \in [n]$}
                \FOR{$j \in [q]$}
                \STATE $\lambda_{i,j}^{k} \gets \lambda_{i,j}^{k-1} + \rho_{i, j}^{k-1} \cdot \zeta( \bm{\delta}^{k}, \lambda_{i,j}^{k-1}, \rho_{i,j}^{k-1})$ \hfill \textit{// dual variables $\lambda_{i,j}$ update}
                \ENDFOR
                
                \STATE $\mu_i^{k} \gets \mu_i^{k-1} + \varrho_i^{k-1} \cdot \xi( \bm{\delta}^{k}, \mu_i^{k-1}, \varrho_i^{k-1}) $ \hfill\textit{// dual variables $\mu_i$ update}
                
            \IF{$k \% r == 0$}
                \FOR{$j \in [q]$}
                    \IF{$g_j(\bm{x}_i + \bm{\delta}^k) > b_j $}
                    \STATE $\rho_{i, j}^k \gets \min (\Bar{\upsilon}, \ \tau \cdot \rho_{i, j}^{k-1}) $ \hfill \textit{// penalty variables $\rho_{i,j}$ update}\\
                    \ENDIF
                \ENDFOR
                
                \IF{$ y\cdot f(\bm{x}_i + \bm{\delta}^k) > c $}
                    \STATE $\varrho_i^k \gets \min(\Bar{\upsilon}, \ \tau \cdot \varrho_i^{k-1}) $ \hfill \textit{// penalty variables $\varrho_{i}$ update}
                \ENDIF
                
            \ENDIF
        \ENDFOR
        
    \ENDWHILE
\end{algorithmic}
\noindent\textbf{Output:} Perturbation $\bm{\delta}_{\bm{X}} \!=\! \argmax_{k \in [K]}\ASR(\bm{\delta}^{k})$
\end{algorithm}

\noindent Algorithm~\ref{alg:CAPX} outlines the \capX{} method, which solves the constrained optimization problem in \eqref{eq:X_final_uncon_UAP_obj} to compute a candidate universal perturbation. The algorithm is iterative, with the number of iterations controlled by a parameter $K$. The learning rate $\alpha$ governs the update magnitude for the primal variable $\bm{\delta}$, while the penalty scaling factor $\tau$ controls the adjustment of penalty terms used to enforce constraint satisfaction. To promote stable convergence and effective constraint enforcement, we initialize the dual variables from a uniform distribution $\mathcal{U}[0, 10^{-4}]$, which is standard practice. Penalty parameters are initialized to 1, providing a balanced scale—sufficient to influence optimization dynamics without introducing instability. Both $\alpha$ and $\tau$ are hyperparameters and can be tuned via cross-validation. 

Algorithm~\ref{alg:CAPX} presents an iterative procedure for learning a universal perturbation $\bm{\delta}_{\bm{X}}$ for a given set $\bm{X}$. It solves the min-max optimization problem in~\eqref{eq:X_final_uncon_UAP_obj} by optimizing
the \emph{primal variable} $\bm{\delta}$ and \emph{dual variables} $\{\lambda_{i,j}\}_{i,j},\{\mu_i\}_{i}$ alternatively. The primal update is performed via gradient descent scaled by learning rate $\alpha$, where dual variables are updated via gradient ascent scaled by penalty variables $\{\rho_{i,j}\}$ and $\{\varrho_i\}$ respectively. The penalty update is done by a multiplicative scale factor $\tau > 1$. We update $\varrho_i$ only when the perturbation to $\bm{x}_i$ fails to induce missclassification, and $\rho_{i,j}$ only when $j$-th constraint gets violated by the perturbed $\bm{x}_i$. Since $\tau > 1$, the penalty parameters always increase at each update, ensuring increasingly severe penalties for constraint violations and missclassifications after each primal update. While penalizing constraint violations and missclassifications is desirable, excessively aggressive penalties can destabilize the optimization process. To mitigate this, we introduce a control factor $r \in [1, K]$ that regulates the update frequency. Moreover, an upper bound $\Bar{\upsilon}$ (e.g., $10^4$) is imposed on the penalty parameters to prevent them from becoming too large to make the objective unbounded. 

Finally, \capX uses the attack success rate ($\ASR$) over the training set $\bm{X}$ to select the best-performing universal perturbation across the iterations. The perturbation that achieves the highest $\ASR$ is returned after $K$ iterations and is expected to generalize well to unseen test examples.

\paragraph{Practical Considerations} The for-loop (lines $3-18$) over the training set $\bm{x}_i, i \in [n]$, and within this, the for-loops (lines $4-6$ and lines $9-13$) over the constraints $g_j, j \in [q]$, can be fully parallelized to update dual variables $\lambda_{i,j}, \mu_i$, and penalty parameters $\rho_{i,j}, \varrho_i$, resulting in substantial reductions in runtime. Moreover, to ensure comparability with prior norm-constrained methods, \capX can be modified to perform a projection onto the $\ell_2$-ball of radius $\epsilon$ whenever the perturbation yields $\norm{\bm{\delta}^k}_2 > \epsilon$. The projected perturbation can be used for evaluation purposes without interfering with the internal
optimization process.

\section{Experimental Results}
All experiments were conducted on an Ubuntu 24.04 Server LTS system with an AMD Ryzen Threadripper PRO 5945WX CPU, 64 GB DDR4 RAM (3200 MT/s), and an NVIDIA RTX A4500 GPU (20 GB GDDR6). Implementations were based on PyTorch and related open-source Python libraries.
For baseline comparisons, we used source code where available, applying necessary modifications to ensure compatibility with our experimental framework.

\begin{wrapfigure}{r}{0.5\textwidth}
    \vspace{-40pt}
    \makebox[\linewidth]{%
        \begin{minipage}[t]{0.48\linewidth}
            \centering LCLD Dataset
        \end{minipage}%
        \hfill
        \begin{minipage}[t]{0.48\linewidth}
            \centering IoMT Dataset
        \end{minipage}%
    }
    \begin{subfigure}[t]{\linewidth}
        \includegraphics[width=0.48\linewidth]{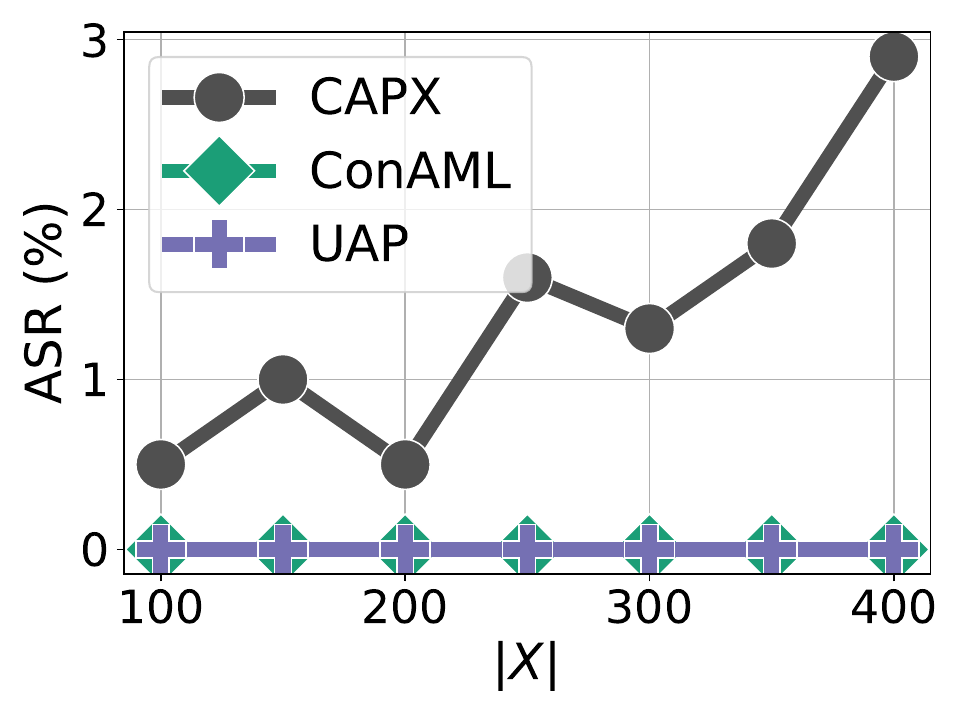}
        \includegraphics[width=0.48\linewidth]{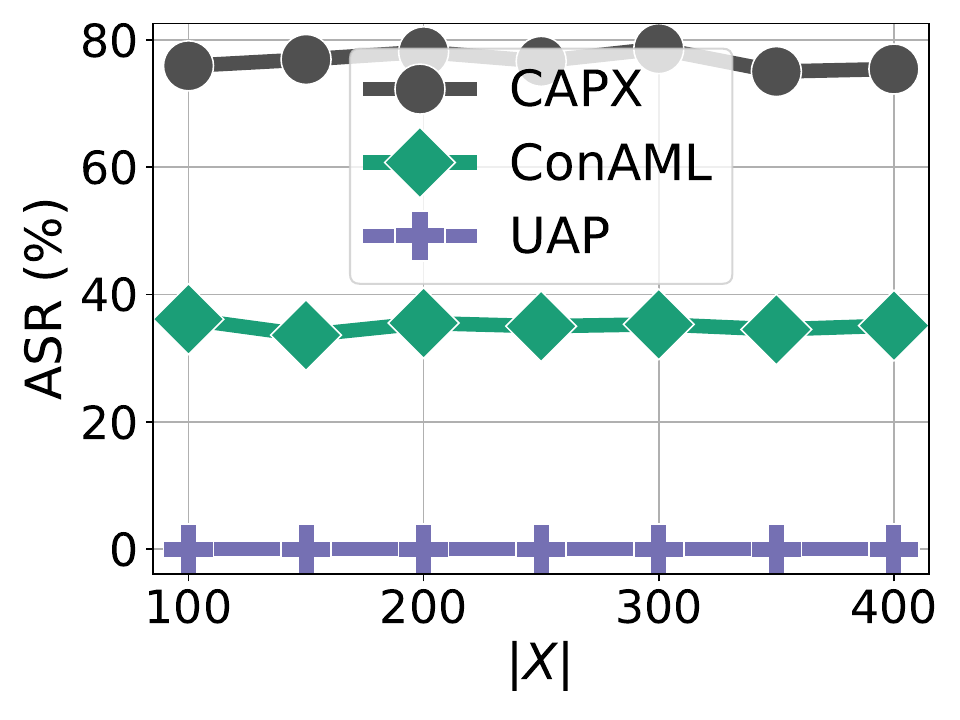}
        \caption{\textit{ASR \ vs.} \ $|\bm X|$}
        \label{fig:LCLD_IoMT_asr_vs_T}
    \end{subfigure}
    \begin{subfigure}[t]{\linewidth}
        \includegraphics[width=0.48\linewidth]{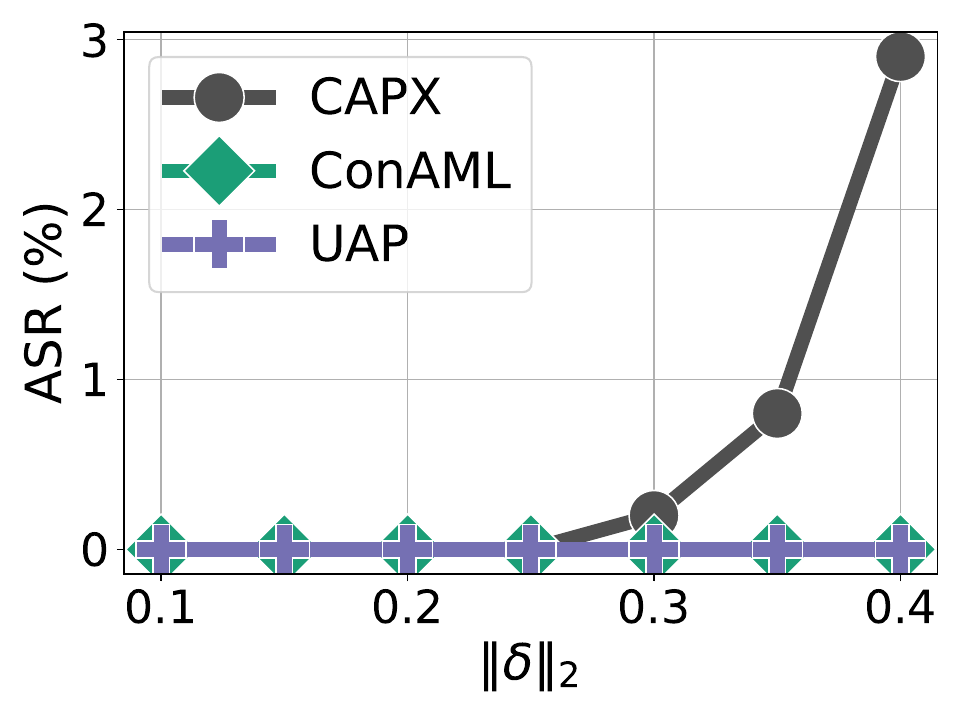}
        \includegraphics[width=0.48\linewidth]{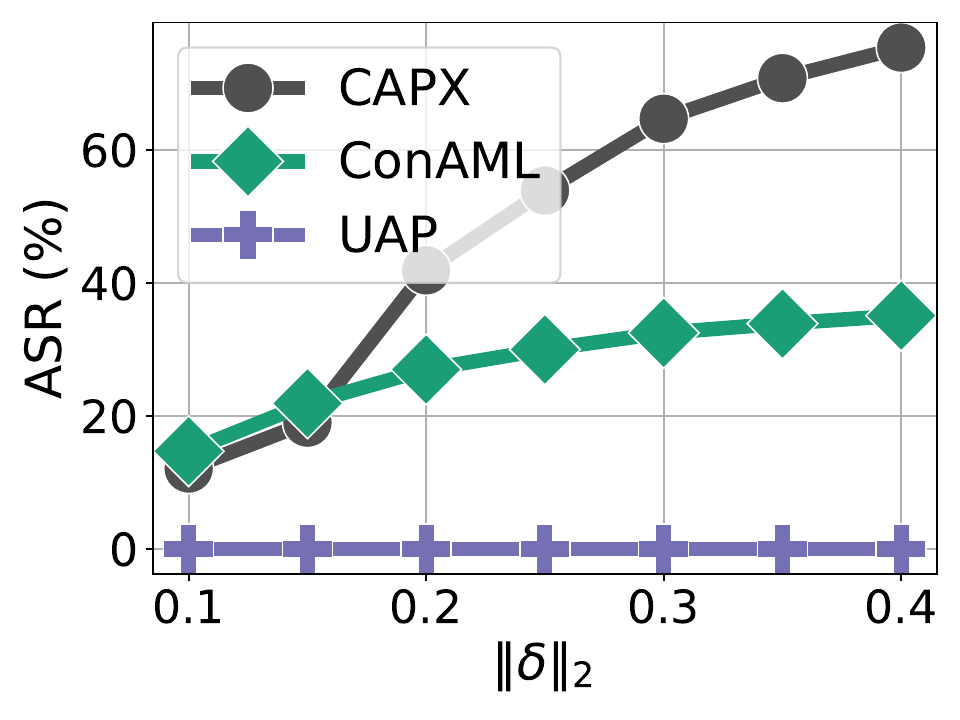}
        \caption{ \textit{ASR \ vs.} \ $\|\bm{\delta}\|_2$}
        \label{fig:LCLD_IoMT_asr_vs_l2}
    \end{subfigure}
    \begin{subfigure}[t]{\linewidth}
        \includegraphics[width=0.48\linewidth]{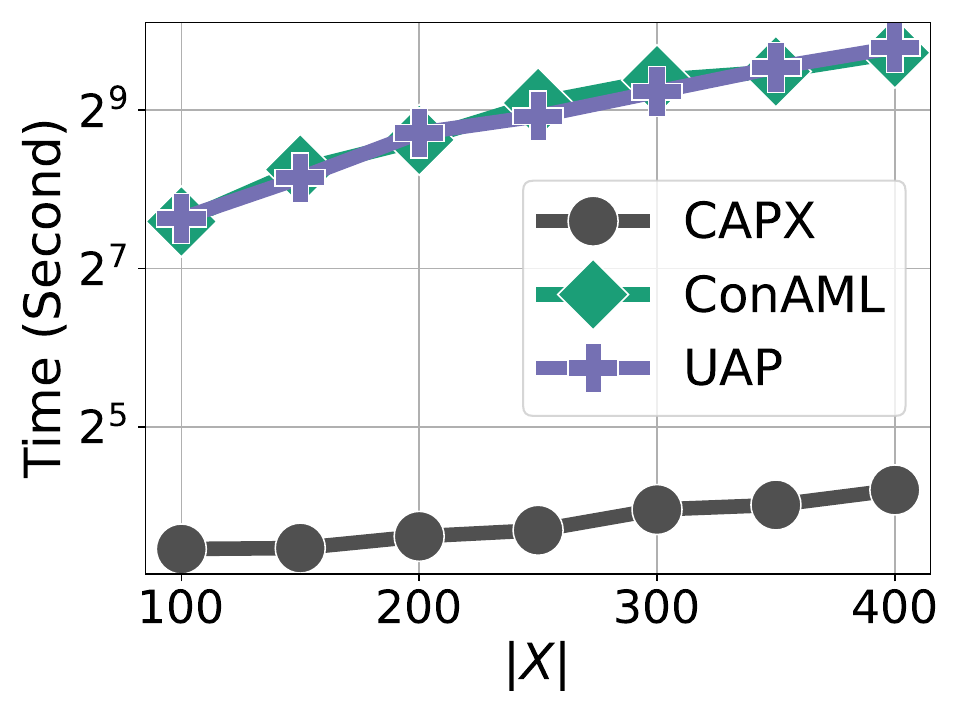}
        \includegraphics[width=0.48\linewidth]{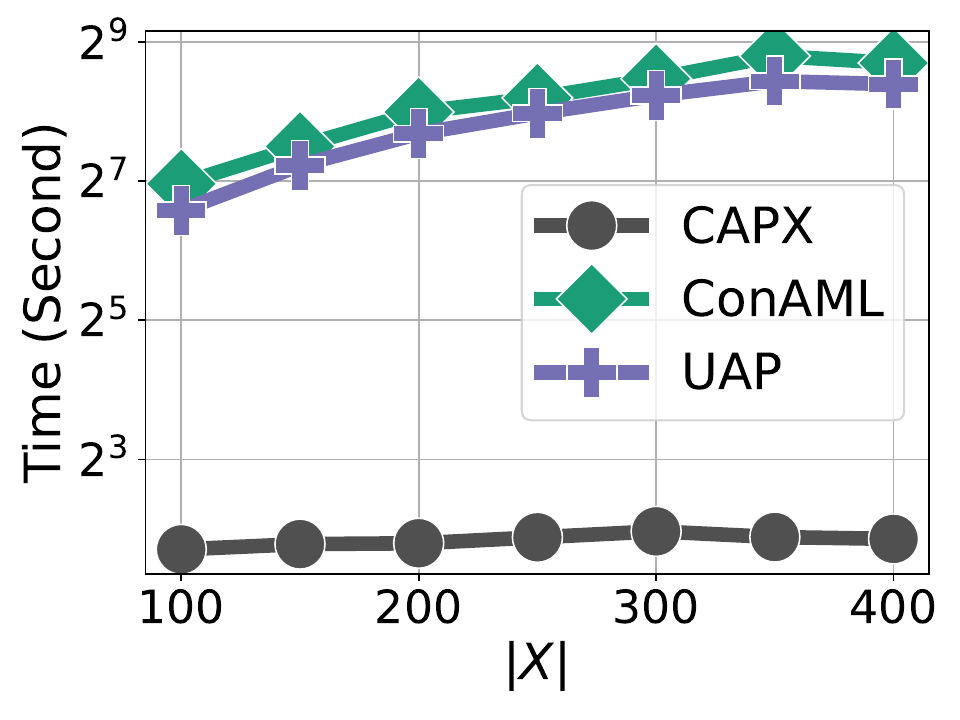}
        \caption{\textit{Time \ vs.} \ $|\bm X|$}
        \label{fig:LCLD_IoMT_time_vs_X}
    \end{subfigure}
    \caption{
        The dependence of adversarial efficacy of \capX and baselines over the size of $\bm{X}$ and the $\ell_2$-norm budget is observed. The left columns correspond to the evaluation result over the test set for the LCLD dataset, and the right columns to the IoMT.  Subplots~\eqref{fig:LCLD_IoMT_asr_vs_T} observe the effect of size $|\bm{X}|$, \eqref{fig:LCLD_IoMT_asr_vs_l2} observe the effect of attack radius of perturbation $\norm{\bm{\delta}}_2$ over the adversarial efficacy of the universal perturbation. Subplots \eqref{fig:LCLD_IoMT_time_vs_X} observe the resultant computational cost. 
        }
    \label{fig:X_LCLD_IoMT_ASR_size_epsilon_time}
    \vspace{-50pt}
\end{wrapfigure}

\paragraph{Datasets}
We evaluate our method across five real-world datasets spanning diverse domains. LCLD \cite{george2020lending} is the Lending Club Loan approval dataset obtained from Kaggle and is used in the work \cite{simonetto2021unified}. IDS \cite{sharafaldin2018toward} is a network traffic dataset, containing both normal and malicious activities for intrusion detection. IoMT \cite{dadkhah2024ciciomt2024} is a healthcare dataset capturing benign and malicious traffic from 40 IoMT devices across protocols such as Wi-Fi, Bluetooth, and MQTT. From the cyber-physical systems domain, SWaT \cite{goh2017dataset} and WADI \cite{ahmed2017wadi} correspond to water treatment and water distribution plants, respectively. Along with the quantitative summary provided in Table~\ref{tab:dataset_n_clf}, a detailed dataset descriptions are included in the appendix.

\paragraph{Classifier}
We use one-dimensional Convolutional Neural Networks (1D-CNN) over the LCLD dataset and a feed-forward neural network (FFNN) for other datasets---both standard choices for structured data.
Models use ReLU activations and SGD optimization. Details on performance, architecture, and parameters are in the appendix.

\subsection{Evaluation of \capX}
\paragraph{Baselines}
Performance of the proposed \capX method is empirically compared with baselines: UAP and ConAML; both employ a three-level nested loop structure. Let $s$ denote the number of outer iterations---controls the maximum allowed iterations, $n$ denotes the intermediate iteration size---used to traverse each $\bm{x}_i \in \bm{X}$, and $t$ the inner most iteration size---the maximum number of iterations allowed for the DeepFool (a method that UAP uses as sub-routine to perturb input $\bm{x}_i \in \bm{X}$ and similarly ConAML employs similar method inspired by DeepFool). In addition to inheriting the computational burden of UAP, ConAML introduces an additional projection step to enforce linear constraints on the feature space, further increasing its per-iteration computational cost compared to UAP. For a fair comparison, we fix the outermost loop $s=30$ and the innermost loop $t=20$ iterations. The intermediate loop for UAP and ConAML scales with the size of the training set $\bm{X}$. In contrast, \capX adopts a single-loop architecture, and is set to $ \ s\times t$ iterations to match the total computational budget of the baselines.

\paragraph{Evaluation} 
The set X, used to learn the universal perturbation for each method, consists of correctly classified normal samples randomly selected from the validation set. A separate set of 3,000 correctly classified normal samples is randomly selected from the test set and used as unseen data to test the generalization ability of perturbation. Table~\ref{tab:X_table} summarizes the adversarial efficacy of \capX and baselines, where \capX consistently achieves higher ASR while requiring significantly lower computation time.

\begin{wrapfigure}{r}{0.5\textwidth}
    \centering
    \begin{footnotesize}
    \begin{tabular}{l@{}r|r|r|r}
    \toprule

    {Dataset \ } & {\ Method} & \textit{$\text{ASR}_{\text{Train}}\uparrow$} & \textit{$\text{ASR}_{\text{Test}}$ $\uparrow$} & \textit{Time (s)$\downarrow$}\\
    
    \midrule
    \multirow{3}{*}{LCLD}
    &\textbf{\capX}&\textbf{3.75}& \textbf{2.90}& \textbf{18.50}  \\     
    &{ConAML}      &   0.00       &    0.00      &   843.26\\
    &{UAP}         &   0.00       &    0.00      &   882.73\\       
                
    \midrule

    \multirow{3}{*}{IDS}
    &{\textbf{\capX}}& \textbf{62.00} & \textbf{58.30} & \textbf{7.27}\\
    &{ConAML}          & 8.75          & 10.30          &   414.50\\
    &{UAP}             & 5.50           & 4.40           & 388.23 \\

    \midrule 
    
    \multirow{3}{*}{IoMT}
    &{\textbf{\capX}}&\textbf{77.00}&\textbf{75.40}&\textbf{3.62}\\       
    &{ConAML}            & 38.00        & 35.10        &  416.35\\
    &{UAP}         & 0.00         & 0.00         & 336.97 \\     
    
    \midrule 
    
    \multirow{3}{*}{SWaT}
    & {\textbf{\capX}} &\textbf{50.25}&\textbf{49.60}&  \textbf{2.62}\\
    & {ConAML}         &12.25         &10.70         &458.11\\
    & {UAP}            &0.00           &0.00          &283.58\\     
    
    \midrule
    
    \multirow{3}{*}{WADI}
    & {\textbf{\capX}}&\textbf{12.50}&\textbf{12.20}&\textbf{3.27}\\
    & {ConAML}        & 0.25        & 1.10         & 415.08\\
    & {UAP}           & 0.00         &0.00          &395.02 \\   

    \bottomrule
  \end{tabular}
  \end{footnotesize}
      \caption{
    \textit{$\text{ASR}_\text{Train}$} and \textit{$\text{ASR}_\text{Test}$} are ASRs of the universal perturbation over the set $\bm{X}$ itself and unseen test examples, respectively. \textit{Time (s)} reports total computation cost in seconds.  \capX offers superior adversarial efficacy and requires significantly lower computation time over baselines}
    \label{tab:X_table}
    \vspace{-10pt}
\end{wrapfigure}

\textit{ASR vs. $|\bm{X}|$: } 
The dependence of ASR of universal perturbation over $|\bm{X}|$ is observed by varying it from 100 to 400 in increments of 50, keeping the perturbation radius fixed at $\epsilon = 0.4$, across all the datasets.  Figure~\ref{fig:LCLD_IoMT_asr_vs_T} observes the dependence of ASR over the size of the set $\bm{X}$ using the test set from the LCLD and IoMT datasets. On the IoMT dataset, increasing the size of the set $\bm{X}$ has minimal impact on the attack generalization of the universal perturbation to unseen examples. Similar trends are observed across the IDS, SWaT, and WADI datasets---Figure~\ref{fig:app_X_asr_vs_X} in the appendix. Over the LCLD dataset, the generalization improves with the inclusion of more samples in $\bm{X}$, attributing the generalization of the universal perturbation to samples in $\bm{X}$.

\textit{ASR vs. $\|\bm \delta\|_2$: } 
Naturally, the attack radius significantly influences ASR, with larger budgets yielding higher success rates. To verify it, we fix the training set size $|\bm{X}| = 400$ and vary the $\ell_2$-norm of the perturbation $\bm{\delta}$ from 0.10 to 0.40 in steps of 0.05, observing the resulting ASR. Figure~\ref{fig:LCLD_IoMT_asr_vs_l2} presents the results over the LCLD and IoMT datasets, while Figure~\ref{fig:app_X_asr_vs_l2} in the appendix reports results over the remaining datasets along with observation of the ASR over the set $\bm{X}$. The results show that \capX achieves significantly higher ASR over both $\bm{X}$ and the test set as the attack radius increases, leveraging the advantage of the systematic handling of the constraints.

\begin{wrapfigure}{r}{0.55\textwidth}
    \vspace{-20pt}
    \centering
    \begin{footnotesize}
    \begin{tabular}{l|rr|rr}
        \toprule
        \multirow{2}{*}{Dataset} &
        \multicolumn{2}{c|}{Training dataset} &
        \multicolumn{2}{c}{Test dataset}\\
        \cmidrule(lr){2-3} \cmidrule(lr){4-5}
        & \textit{$\text{IP}_{\text{Normal}}\downarrow$} & \textit{$\text{IP}_\text{Anomal.}\uparrow$} & \textit{$\text{IP}_\text{Normal}\downarrow$} & \textit{$\text{IP}_\text{Anomal.}\uparrow$}\\
        \midrule
        LCLD    &   0.0033  &   0.0033  &   0.0033  &   0.0033 \\
         IDS    &   0.0036  &   0.0056  &   0.0036  &   0.0055 \\
        IoMT    &   0.0035  &   0.1311  &   0.0034  &   0.1310 \\
        SWaT    &   0.0098  &   0.1658  &   0.0098  &   0.1687 \\
        WADI    &   0.0098  &   0.2313  &   0.0097  &   0.2327 \\
        \bottomrule
    \end{tabular}
    \end{footnotesize} 
       \caption{Summary of average inner product (IP) of a null space vector $\bm{c}$ taken with the normal samples (\textit{$\text{IP}_\text{Normal}$}) and anomalous samples \textit{($\text{IP}_\text{Anomal.}$)} over the training and the test dataset. 
       IP is almost preserved for the normal readings of the test set, signifying its importance as a constraint on the normal readings. On the other hand, the inner product magnitudes rise for anomalous readings across datasets.}
    \label{tab:inner_product}
    \vspace{-20pt}
\end{wrapfigure}

\textit{Time vs. $|\bm X|$:} 
While observing the effect of the size of the set $\bm{X}$ on the ASR as reported in Figure~\ref{fig:LCLD_IoMT_asr_vs_T}, the total runtime of each method was observed with $|\bm{X}|$ as well. As shown in Figure~\ref{fig:LCLD_IoMT_time_vs_X}, the runtime of UAP and ConAML varies significantly with the size of $\bm{X}$ as both methods traverse over each element in the set $\bm{X}$ as a subroutine to learn the universal perturbation, whereas \capX exhibits a much slower growth in computation time as it uses $\bm{X}$ to calculate the universal objective \eqref{eq:X_simplified_ALF} that can be readily parallelized to avoid intermediate loops. Similar trends are observed across other datasets, as reported in Figure~\ref{fig:asr_vs_time} in the appendix.

\begin{wrapfigure}{r}{0.58\textwidth}
    \vspace{-30pt}
    \centering
    \begin{footnotesize}
    \begin{tabular}{r@{}l|r|r|r|r|r}
    \toprule
    \multicolumn{2}{r|}{Method} & {LCLD} & {IDS} & {IoMT} & {SWaT} &  {WADI}\\

    \midrule 
    \multirow{3}{*}{\textit{*ASR$\uparrow \ \ $}}
    & \textbf{\capx}&\textbf{45.2}&\textbf{83.8}&\textbf{97.8}&\textbf{91.2}&\textbf{25.6}\\
    & {MoEv.} &19.4  & 63.4        & 95.2        & 89.0        &  3.8\\
    & {PGD}  & 0.0  & 2.2        & 12.6        & 32.8        &  1.0\\

    \midrule

    \multirow{3}{*}{\textit{*T-ms$\downarrow \ \ $}}
    & \textbf{\capx} & 322.7  &    7.0 &       7.0 &       6.1  &       5.7 \\
    & {MoEv.} & 367.7 &   120.2&        124.5&        117.7 &  118.2\\  
    & {PGD}   & \textbf{177.5} &\textbf{2.7}&\textbf{2.8}&\textbf{2.7}&\textbf{2.4}\\

    \midrule
    \multirow{2}{*}{\textit{$\circ$ASR$\uparrow \ \ $}}
    & \textbf{\capx}&56.6& 94.4 &\textbf{97.4}&\textbf{90.4}&\textbf{28.8}\\
    & {MoEv.} &\textbf{69.4}  & \textbf{95.6}        & 97.0        & 90.0        &  28.0\\

    \midrule

    \multirow{2}{*}{\textit{$\circ$T-ms$\downarrow \ \ $}}
    & \textbf{\capx} & \textbf{325.6}  &    \textbf{7.6} &       \textbf{6.9} &       \textbf{5.3}  &       \textbf{5.8} \\
    & {MoEv.} & 2030.0 &   554.2&        589.3&        571.5 &  573.5\\

    \bottomrule
  \end{tabular}
  \end{footnotesize}
      \caption{Summary of adversarial efficacy of individual perturbation when norm budget over $\|\bm{\delta}\|_2$ is fixed to 0.4.
      In addition to the maximum allowed iteration---common to each method, MoEvA2 includes two more parameters, namely population and offspring size.  
      \textit{*T-ms} and \textit{$\circ$T-ms} report time in milliseconds, and they represent the average computational time needed by methods in each step of the iteration. 
      Rows in \textit{*ASR} report ASR when maximum allowed iteration is fixed to 300 for each method (MoEvA2: population size fixed to 64 and offspring size to 32), and rows in \textit{*T-ms} report the corresponding average per-step time.
      \capx and MoEvA2 are more direct competitors as they both perturb under the feature constraints, unlike PGD.
      Rows in \textit{$\circ$ASR} report ASR when the maximum allowed iteration is increased to 625 for both methods (MoEvA2: population size increased to 640 and offspring size increased to 320), and rows in \textit{$\circ$T-ms} report the corresponding time.
      Despite operating under the same iteration budget, \capx, achieves competitive ASR performance and greater time efficiency even when MoEvA2 is configured with larger population and offspring sizes that typically benefit genetic algorithms.
      }
    \label{tab:x_table}
\end{wrapfigure}

\subsection{Constraint Learning from Data}
After having the vector $\bm{c}$ (say sum of the basis vectors of the null space), we empirically validate it by comparing the average inner products $\bm{x}^\top \bm{c}$ for normal and anomalous examples, $\bm x$, across both training and test sets (Table~\ref{tab:inner_product}). The results show that the inner product of $\bm{c}$ with normal examples yields consistently low values (on the order of $10^{-2}$), while anomalous examples produce significantly larger values. Furthermore, Figure~\ref{fig:data_rank}(a) illustrates that the nullity of $\bm{X}_+$ remains positive even as more normal training samples are added, indicating the presence of linear dependency in features, across all datasets.

In domains like cyber-physical systems, identifying invariants that capture physical dependencies across sensors and actuators becomes increasingly challenging as system complexity grows \cite{adepu2016using, quinonez2020savior}. To address this, we propose a data-driven method for extracting linear feature constraints by analyzing the null space of training examples from the normal class $\bm{X}_+ \!\in\! \mathbb{R}^{n \times d}$. Specifically, we compute a non-zero vector $\bm{c}$ in the null space such that $\bm{X}_+ \bm{c} \!=\! 0$, implying a normal sample $\bm{x}$ should approximately satisfy $\bm{x}^\top \bm{c} \approx 0$.

In our experiments, we treat these null-space relations as linear constraints. Perturbed samples $\bm{x} + \bm{\delta}$ are required to satisfy $(\bm{x} + \bm{\delta})^\top \bm{c} \le \text{IP}_\text{mean} + \text{IP}_\text{std.}$, where $\text{IP}_\text{mean}$ and $\text{IP}_\text{std.}$ are the empirical mean and standard deviation of inner products computed over the normal training set. We augment the known constraint of the IoMT dataset with a null-space constraint. For SWaT and WADI, we rely solely on the learned null-space constraints, and for LCLD and IDS, we only use known constraints over the dataset.

\section{Individual Perturbation: \capx}
The universal perturbation method proposed in Section \ref{sec:proposed_method} can be directly adapted to individual perturbations under feature constraints---we refer to the method as \capx. The derivation and corresponding algorithm are provided in the appendix for reference. Below, we directly present the optimization formulation corresponding to the individual perturbation problem \eqref{eq:cap}, as introduced in Section \ref{sec:background_and_problem_setup}:
\begin{align*}
  \min_{\bm{\delta}} \ \max_{\bm{\lambda}, \mu} \  \mathcal{L}( \bm{\delta}, \bm{\lambda}, \mu, \bm{\rho}, \varrho) = &  \ \frac{1}{2}\|\bm{\delta}\|^2 
        + \sum\nolimits_{j=1}^q \lambda_j \cdot \zeta(\bm{\delta}, \lambda_j, \rho_j)
        \ + \ \mu \cdot \xi(\bm{\delta}, \mu, \varrho)  \\
        + & \frac{1}{2} \sum\nolimits_{j=1}^q \rho_j \cdot \zeta_j(\bm{\delta}, \lambda_j, \rho_j)^2 
        \ + \ \frac{1}{2} \varrho \cdot \xi(\bm{\delta}, \mu, \varrho)^2,
\end{align*}
where $\bm{\delta} \in \mathbb{R}^d$ is the primal variable, $\bm{\lambda} \in \mathbb{R}^q, \mu \in \mathbb{R}$ are the dual variables, $\bm{\rho} \in \mathbb{R}^q, \varrho \in \mathbb{R}$ are the penalty variables, $q$ is the number of feature constraints, and 
\begin{subequations}
    \begin{align*}
        \zeta(\bm{\delta}, \lambda_j, \rho_j) &:= \max \lbrace g_j(\bm{x} + \bm{\delta}) - b_j, \ -\lambda_j/\rho_j \rbrace, \\
        \xi(\bm{\delta}, \mu, \varrho) &:= \max \lbrace y\cdot f(\bm{x} + \bm{\delta}) - c, \ -\mu/\varrho \rbrace~.
    \end{align*}
\end{subequations}

All methods were evaluated on CPU cores to ensure a fair comparison, as MoEvA2 lacks support for parallelization. Despite this constraint, \capx demonstrates strong performance—outperforming existing methods under low iteration budgets and remaining competitive as the budget increases, as shown in Table~\ref{tab:x_table}. Unlike MoEvA2, \capx is designed to be parallelizable, enabling efficient GPU utilization. This reduces computational time and enhances its suitability for real-time adversarial scenarios.

\section{Conclusion}
The present work proposes a systematic method to find individual and universal adversarial perturbations under feature constraints. While the domain-specific feature constraints offer significant resistance against the adversarial methods that do not consider feature constraints. However, the attack superiority of the proposed method \capX and \capx and the computational efficiency expose the vulnerabilities of the current system against online and offline attacks against individual and universal perturbation. This study highlights the need for future studies on the robustness guarantee against adversarial attacks, given that the adversary is restricted to perturb under the feature constraints.

\clearpage
\bibliographystyle{plainnat}
\bibliography{ref}

\appendix
\section{Appendix}
\subsection{Proof of Lemma \ref{le:dual_gradient}}
Dual variables are mutually independent, and the terms involving them in the universal adversarial problem \eqref{eq:X_final_uncon_UAP_obj} have a similar structure. Hence, in order to obtain the gradient of the objective \eqref{eq:X_simplified_ALF} with respect to the dual variable, it will be sufficient to compute the partial derivative with respect to an arbitrary dual variable, say $\lambda_{i,j}$, and infer the derivatives with respect to the rest of the dual variables. In this regard, taking the partial derivative of $\mathcal{L}(\bm{\delta}, \bm{\Lambda}, \bm{\mu}, \bm{P}, \bm{p})$---defined as objective \eqref{eq:X_simplified_ALF},  with respect to $\lambda_{i,j}$ yields the following: 
\begin{align} \label{eq:app_first_part_der}
    \partial_{\lambda_{i,j}}\mathcal{L}(\bm{\delta}, \bm{\Lambda}, \bm{\mu}, \bm{P}, \bm{p}) &= \ \partial_{\lambda_{i,j}} \sum\nolimits_{i=1}^{n}\sum\nolimits_{j=1}^q \lambda_{i,j} \cdot \zeta(\bm{\delta}, \lambda_{i, j}, \rho_{i, j}) \\
    +& \ \partial_{\lambda_{i,j}} \frac{1}{2}\sum\nolimits_{i=1}^{n}\sum\nolimits_{j=1}^q \rho_{i, j} \cdot \zeta(\bm{\delta}, \lambda_{i, j}, \rho_{i, j})^{2}.  \nonumber
\end{align}

\noindent Furthermore considering only the terms in \eqref{eq:app_first_part_der} dependent on $\lambda_{i,j}$ yields the following:
\begin{align} \label{eq:app_derivative_intermediate_res}
    \partial_{\lambda_{i,j}} \mathcal{L}(\bm{\delta}, \bm{\Lambda}, \bm{\mu}, \bm{P}, \bm{p})
    =& \ \lbrace \partial_{\lambda_{i,j}} \lambda_{i, j}\rbrace \cdot \zeta(\bm{\delta}, \lambda_{i, j}, \rho_{i,j})
    + \ \lambda_{i, j} \cdot \lbrace\partial_{\lambda_{i,j}} \zeta(\bm{\delta}, \lambda_{i, j}, \rho_{i, j})\rbrace \\ 
    +& \ \rho_{i, j} \cdot \zeta(\bm{\delta}, \lambda_{i, j}, \rho_{i, j}) \cdot \partial_{\lambda_{i,j}} \lbrace\zeta(\bm{\delta}, \lambda_{i, j}, \rho_{i, j}) \rbrace. \nonumber
\end{align}

\noindent To proceed further, the expression for the partial derivative of $\zeta(\bm{\delta}, \lambda_{i, j}, \rho_{i, j})$ with respect to $\lambda_{i, j}$ is needed. As per the definition in \eqref{eq:def_of_zeta_n_xi}---in the main article, we have:
\begin{align}
    \zeta(\bm{\delta}, & \lambda_{i, j}, \rho_{i, j}) = \max\left \lbrace g(\bm{x}_i + \bm{\delta}) - b_j, \ -\lambda_{i, j} / \rho_{i, j} \right \rbrace.\nonumber
\end{align}

\noindent Since, $\zeta(\bm{\delta}, \lambda_{i, j}, \rho_{i, j})$ is a $\max$ function defined over two quantities, hence the partial derivative $\partial_{\lambda_{i,j}} \zeta(\bm{\delta}, \lambda_{i, j}, \rho_{i, j})$ depends over the following two cases:
\begin{align*}
        g(\bm{x}_i + \bm{\delta}) - b_j <& \ -\lambda_{i, j} / \rho_{i, j}~,\\
        g(\bm{x}_i + \bm{\delta}) - b_j \geq & \ -\lambda_{i, j} / \rho_{i, j}~.
\end{align*}
\noindent Consideration of the previous two cases---as part of definition of $\zeta(\bm{\delta}, \lambda_{i, j}, \rho_{i, j})$, yields the following:
\begin{align} \label{eq:app_partial_derivative_res}
    \partial_{\lambda_{i,j}} \zeta(& \bm{\delta},\lambda_{i, j}, \rho_{i, j}) \\
    = & \begin{cases}
        -1/\rho_{i,j} &\text{if $g_j(\bm{x}_i + \bm{\delta}) - b_j < -\lambda_{i, j}$} / \rho_{i, j}~, \nonumber\\
        \ \ \ 0  &\text{otherwise}~. \nonumber
    \end{cases}
\end{align}
\noindent On substituting the result obtained from \eqref{eq:app_partial_derivative_res} into \eqref{eq:app_derivative_intermediate_res} we obtain:
\begin{align} \label{eq:final_partial_derv_expresssion_app}
    & \partial_{\lambda_{i,j}} \mathcal{L}(\bm{\delta}, \bm{\Lambda}, \bm{\mu}, \bm{P}, \bm{p)} \\
    = & \begin{cases}
        -\lambda_{i,j} / \rho_{i,j}&\text{if $g_j(\bm{x}_i + \bm{\delta}) - b_j < \ -\lambda_{i,j}/\rho_{i, j}$}~,\\
        \zeta(\bm{\delta}, \lambda_{i, j}, \rho_{i, j})  &\text{otherwise}~.
    \end{cases} \nonumber
\end{align}

\noindent By using definition of $\zeta(\bm{\delta}, \lambda_{i, j}, \rho_{i, j})$, we can write
\begin{align} \label{eq:app_final_part_derivative}
    \partial_{\lambda}& {}_{{}_{i,j}} \mathcal{L}(\bm{\delta}, \bm{\Lambda}, \bm{\mu}, \bm{P}, \bm{p}) \nonumber \\
    = & \ \begin{cases}
        -\lambda_{i,j} / \rho_{i,j} &\text{if $g_{j}(\bm{x}_i + \bm{\delta}) - b_{j} < \ -\lambda_{i,j} /\rho_{i,j}$}\\
        g_j(\bm{x}_i + \bm{\delta}) - b_j  &\text{otherwise}
    \end{cases} \nonumber \\
    = & \ \max\left \lbrace g_j(\bm{x}_i + \bm{\delta}) - b_j, \ -\lambda_{i,j} /\rho_{i,j} \right \rbrace \nonumber \\
    = & \ \ \zeta(\bm{\delta}, \lambda_{i, j}, \rho_{i, j})~.
\end{align}

The equation \eqref{eq:app_final_part_derivative} gives the partial derivative of the universal perturbation objective \eqref{eq:X_final_uncon_UAP_obj} with respect to the dual variables $\lambda_{i,j}$. Owing to the independence of the dual variables and similarities of the expression involving them in the objective \eqref{eq:X_simplified_ALF}, the following can be concluded:
\begin{subequations} \label{eq:gradient_proof}
    \begin{align}
        \partial_{\lambda_{i, j}} \mathcal{L}(\bm{\delta}, \bm{\Lambda}, \bm{\mu}, \bm{P}, \bm{p}) &= \zeta(\bm{\delta}, \lambda_{i, j}, \rho_{i, j})~,\\
        \partial_{\mu_i} \mathcal{L}(\bm{\delta}, \bm{\Lambda}, \bm{\mu}, \bm{P}, \bm{p}) &= \xi(\bm{\delta}, \mu_i, \varrho_i)~.
    \end{align}
\end{subequations}    

\noindent Since the objectives for both individual and universal perturbations are derived using a similar mathematical framework. Hence, the adversarial objective expression for them exhibits a similar mathematical structure. As a result, the derivatives of the objective function for individual perturbations with respect to the dual variables can be computed in a manner analogous to that of universal perturbation.

\subsection{Individual Perturbation: \capx} \label{ssec:CAPx}
This section provides the development of the proposed methods \capx that solves the individual adversarial problem \ref{eq:cap} discussed in Section \ref{sec:background_and_problem_setup}. To recall, the objective to obtain the individual perturbation that should satisfy the constraint over the feature space was expressed using the following optimization problem:
\begin{subequations} \label{eq:x_main_obj}
    \begin{align}
        \min_{\bm{\delta}} \quad &\frac{1}{2}\|\bm{\delta}\|^2 \label{eq:x_objective}\\
        \text{s.t.}, \quad  & \bm{g}(\bm{x} + \bm{\delta}) \leq \bm{b}~, \label{eq:x_non_lin_cons}\\
         & y \cdot f(\bm{x} + \bm{\delta}) \leq c~, \label{eq:x_mclf_cons}
    \end{align}
\end{subequations}

\noindent where, $c$ is a small negative constant (e.g., $10^{-4}$) used to relax the strict misclassification inequality constraint to a more general non-strict inequality constraint, $\bm{\delta} \in \mathbb{R}^d$ denotes the individual perturbation for the target example $\bm x$ belonging to class $y$, and together the constraint \eqref{eq:x_non_lin_cons}, \eqref{eq:x_mclf_cons} enforce the perturbed example $\bm{x} + \bm{\delta}$ should satisfies the feature constraints while achieving the misclassification.

Since the equality constraints are easier to manage in an optimization framework, hence, we convert all the inequality constraints into equality constraints using slack variables and re-express the individual adversarial problem using the equality constraints as follows:
\begin{subequations} \label{eq:x_obj_eq_cons}
    \begin{align}
        \min_{\bm \delta} \quad & \frac{1}{2}\|\bm{\delta}\|^2\\
        \text{s.t.} \quad  & \bm{g}(\bm{x} + \bm{\delta}) + \bm{\phi} - \bm{b} = 0 \label{eq:x_eq_non_lin_con}~,\\
        & y \cdot f(\bm{x} + \bm{\delta}) + \theta - c = 0 \label{eq:x_eq_mclf_con}~,\\
        & \phi_j \geq 0, \theta \geq 0, \forall j \in [q]~, \label{eq:x_slack_con}
    \end{align}
\end{subequations}

\noindent where, $\bm{\phi} = \lbrace\phi_j\rbrace_{j}, \forall j \in [q]$ and $\theta$ are slack variables with each $\phi_j$ and $\theta \in \mathbb{R}$. The inequalities \eqref{eq:x_slack_con} introduced in the adversarial objective due to the slack variables are simple non-negativity constraints. For now, we will ignore them from the formulation and will address them separately.

The augmented Lagrangian function (ALF) for the optimization problem \eqref{eq:x_obj_eq_cons} is defined as:
\begin{align}
    \mathcal{L}&(\bm{\delta}, \bm{\lambda}, \mu, \bm{\phi}, \theta, \bm{\rho}, \varrho) = \frac{1}{2}\|\bm{\delta}\|^2 \label{eq:x_alf}\\
    & + \sum\nolimits_{j=1}^q \lambda_j \cdot \lbrace g_j(\bm{x} + \bm{\delta}) + \phi_j - b_j \rbrace 
    + \mu \cdot \lbrace y\cdot f(\bm{x} + \bm{\delta}) + \theta - c\rbrace \nonumber\\
    & + \frac{1}{2} \sum\nolimits_{j=1}^q \rho_j \cdot \lbrace g_j(\bm{x} + \bm{\delta}) + \phi_j - b_j \rbrace^2
    + \frac{1}{2} \varrho \cdot \lbrace y \cdot f(\bm{x} + \bm{\delta}) + \theta - c\rbrace^2~.  \nonumber
\end{align}

\noindent Here, $\bm{\lambda} = \{\lambda_j\}_j, \forall j \in [q]$ and $\mu$ are Lagrange multipliers associated with feature and misclassification constraints, respectively, with each $\lambda_j$ and $\mu \in \mathbb{R}$. $\bm{\rho} = \{\rho\}_j, \forall j \in [q]$ and $\varrho$ are the penalty variables associated with feature and misclassification constraints, respectively, with each $\rho_j$ and $\varrho \in \mathbb{R}_+$.

Using the augmented Lagrangian function defined in \eqref{eq:x_alf}, we express the objective to obtain the individual perturbation using the following optimization problem:
\begin{equation}
    \min_{\bm{\delta}, \bm{\phi}, \theta} \ \max_{\bm{\lambda}, \mu} \ 
    \mathcal{L}(\bm{\delta}, \bm{\lambda}, \mu, \bm{\phi}, \theta, \bm{\rho}, \varrho)~.
    \label{eq:x_uncon_obj}
\end{equation}

\noindent The problem \eqref{eq:x_uncon_obj} is quadratic in slack variables, hence using the first-order derivative with respect to $\phi_j$, we obtain \eqref{eq:x_slack_sg_opt} as a condition for optimality:
\begin{align}
    \lambda_j + \rho_j \cdot \lbrace g_j(\bm{x} + \bm{\delta}) + \phi_j - b_j\rbrace = 0 \nonumber\\
    \implies \phi_j = -\lambda_j/\rho_j - g_j(\bm{x} + \bm{\delta}) + b_j 
    \label{eq:x_slack_sg_opt}
\end{align}

\noindent Similarly, equation \eqref{eq:x_slack_sg_sf_opt} serves as the optimality condition for slack variable $\theta$:
\begin{equation} \label{eq:x_slack_sg_sf_opt}
        \theta = -\mu/\varrho - y\cdot f(\bm{x} + \bm{\delta}) + c~.
\end{equation} 

Previously, we ignored the non-negativity constraints on the slack variables over the adversarial formulation. To enforce it, we now apply a projection onto the non-negative orthant, $\Rplus$:
\begin{subequations}
    \begin{align}
        \phi_j \ =& \ \max\left \lbrace 0, \ -\lambda_j/\rho_j - g_j(\bm{x} + \bm{\delta}) + b_j \right \rbrace~, \label{eq:x_slack_sg_opt_exp}\\
        \theta \ =& \ \max\left \lbrace 0, \ -\mu/\varrho - y\cdot f(\bm{x} + \bm{\delta}) + c \right \rbrace. \label{eq:x_slack_sf_opt_exp}
    \end{align}
\end{subequations}

\noindent Using the optimality conditions of the slack variables, the following terms are introduced for notational convenience:
\begin{subequations}
    \begin{align}
        \zeta(\bm{\delta}, \lambda_j, \rho_j) := & \max \lbrace g_j(\bm{x} + \bm{\delta}) - b_j, \ -\lambda_j/\rho_j \rbrace~, \label{eq:x_zeta_g} \\
        \xi(\bm{\delta}, \mu, \varrho) := & \max \lbrace y \cdot f(\bm{x} + \bm{\delta}) - c, \ -\mu/\varrho \rbrace~. \label{eq:x_zeta_f} 
    \end{align}
\end{subequations}

\noindent where, \eqref{eq:x_zeta_g} and \eqref{eq:x_zeta_f} obtained by adding expression $g_j(\bm{x} + \bm{\delta}) - b_j$ and $f(\bm{x} + \bm{\delta}) - c$ to \eqref{eq:x_slack_sg_opt_exp} and \eqref{eq:x_slack_sf_opt_exp} respectively. By incorporating these definitions and the previous simplifications, the augmented Lagrangian function \eqref{eq:x_alf} is simplified as follows:
\begin{align}
    \mathcal{L}(\bm{\delta}, \bm{\lambda}, & \mu, \bm{\rho}, \varrho) =  \ \frac{1}{2}\|\bm{\delta}\|^2 \label{eq:x_final_alf_app}\\
        + & \sum\nolimits_{j=1}^q \lambda_j \cdot \zeta(\bm{\delta}, \lambda_j, \rho_j)
        \ + \ \mu \cdot \xi(\bm{\delta}, \mu, \varrho) \nonumber \\
        + & \frac{1}{2} \sum\nolimits_{j=1}^q \rho_j \cdot \zeta_j(\bm{\delta}, \lambda_j, \rho_j)^2 
        \ + \ \frac{1}{2} \varrho \cdot \xi(\bm{\delta}, \lambda, \rho)^2~, \nonumber
\end{align}
and the adversarial objective for the individual perturbation is simplified as:
\begin{equation}
    \min_{\bm{\delta}} \ \max_{\bm{\lambda}, \mu} \ \mathcal{L}(\bm{\delta}, \bm{\lambda}, \mu, \bm{\rho}, \varrho). \label{eq:x_final_obj_app}
\end{equation}

\begin{algorithm}[t!]
\caption{\textbf{\capx}} \label{alg:CAPx}
\textbf{Input:} Target example $\bm x$, label $y$, classifier $f$, classification threshold $c<0$, constraints $\{g_j, b_j\}_{j=1}^q$, maximum iteration $K$, learning rate $\alpha > 0$, penalty scaling $\tau \geq 1$, penalty frequency update factor $r \leq K$, penalty upper bound $\Bar{\upsilon} > 0$.

\textbf{Variable Initialization:}
Primal: $\bm{\delta}^0 \gets \bm{0}^{d\times 1}$, 
Dual: $\lambda_j^0 \gets \mathcal{U}[0, 10^{-4}], \mu^0 \gets \mathcal{U}[0, 10^{-4}], j \in [q]$, \\
Penalty: $\rho_j^0 \gets 1, \varrho^0 \gets 1, j \in [q]$.

\textbf{Learning steps:}
\begin{algorithmic}[1]
\WHILE{$k \leq K$}
        \STATE $\bm{\delta}^{k} \gets \bm{\delta}^{k-1} - \alpha \nabla_{\bm{\delta}} \mathcal{L}(\bm{\delta}^{k-1}, \bm{\lambda}^{k-1}, \mu^{k-1}, \bm{\rho}^{k-1}, \varrho^{k-1})$ \hfill \textit{// primal variables $\bm{\delta}$ update} \label{alg:delta_update_x}\\
        \textit{// Update dual variable $\lambda_{j}$ and $\mu$}

    \FOR{$j \in [q]$}
        \STATE $\lambda_j^k \gets \lambda_j^{k-1} + \rho^{k-1}\cdot \zeta(\bm{\delta}^k, \lambda_j^{k-1}, \rho_j^{k-1})$ \hfill \textit{// dual variable $\lambda_j$ update}\label{alg:lambda_update_x}
    \ENDFOR
    
        \STATE $\mu^k \gets \mu^k + \mu^k \cdot \xi(\bm{\delta}^{k}, \mu^{k-1}, \varrho^{k-1})$ \hfill \textit{// dual variable $\mu$ update} \label{alg:mu_update_x}\\

    \IF{$k\%r == 0$}
        \FOR{$j \in [q]$}
            \STATE $\rho_j^k \gets \max\{\Bar{\upsilon}, \ \tau\cdot \rho_j^{k-1}\}$ \hfill \textit{// penalty variable $\rho_j$ update}
        \ENDFOR
        
        \STATE $\varrho^{k} \gets \max\{\Bar{\upsilon}, \ \tau\cdot \varrho^{k-1}\}$ \hfill \textit{// penalty variable $\varrho$ update}
        
    \ENDIF
\ENDWHILE
\end{algorithmic}
\paragraph{Output} Perturbation $\bm{\delta}_{\bm{x}} \!=\! \argmax_{k \in [K]}\ASR(\bm{\delta}^{k})$
\end{algorithm}

\subsection{Algorithm: \capx}
Algorithm \ref{alg:CAPx} solves the optimization problem \eqref{eq:x_final_obj_app} to obtain the individual perturbation $\bm{\delta_x}$ for the target example $\bm{x}$ with label $y$. 
The utilities of the primal, dual, and penalty variables introduced for the \capx methods remain similar to the \capX, and similarly, their initialization reason remains the same. Hence, the main article should be referred to for more details regarding the variables introduced for \capx

\begin{table*}[t!]
  \centering
  \begin{scriptsize} 
  \begin{tabular}{lr|rrr|rr|rr|rrrr}
  \toprule
    \multicolumn{2}{c|}{\multirow{2}{*}{Dataset}} & \multicolumn{5}{c|}{Dataset Description} & \multicolumn{6}{c}{Classifiers Performance}\\
    \cmidrule(lr){3-7} \cmidrule(lr){8-13}
    & & Train & Val. & Test & Feat. & $\ell_2$-norm & Accuracy & F1-score & TP & TN & FP & FN\\

  \midrule
    \multirow{2}{*}{LCLD}
    & Rejected & 168471  & 36101  & 36101
    & \multirow{2}{*}{47} &  \multirow{2}{*}{2.52} & \multirow{2}{*}{80.67} & \multirow{2}{*}{89.13} & \multirow{2}{*}{79.20} & \multirow{2}{*}{1.47} & \multirow{2}{*}{18.25} & \multirow{2}{*}{1.07}\\
    & Approved & 685593 & 146913 & 146913 & & & & & & & \\

    \multirow{2}{*}{IDS}
    & Anomal. & 928041  & 198866  & 198866
    & \multirow{2}{*}{76} &  \multirow{2}{*}{2.88} & \multirow{2}{*}{98.08} & \multirow{2}{*}{98.91} & \multirow{2}{*}{87.04} & \multirow{2}{*}{11.04} & \multirow{2}{*}{1.21} & \multirow{2}{*}{0.70}\\
    & Normal & 6645239 & 1423980 & 1423980 & & & & & & & \\

    \multirow{2}{*}{IoMT}
    & Anomal. & 3623581  & 776482  & 776481 
    & \multirow{2}{*}{45} &  \multirow{2}{*}{2.25} & \multirow{2}{*}{99.65} & \multirow{2}{*}{95.93} & \multirow{2}{*}{4.12} & \multirow{2}{*}{95.53} & \multirow{2}{*}{0.21} & \multirow{2}{*}{0.14}\\
    & Normal & 161237 & 34551 & 34551 & & & & & & & \\

    \multirow{2}{*}{SWaT}
    & Anomal. & 38235  & 8193  & 8193 
    & \multirow{2}{*}{51} &  \multirow{2}{*}{4.63} & \multirow{2}{*}{99.25} & \multirow{2}{*}{99.60} & \multirow{2}{*}{94.10} & \multirow{2}{*}{5.15} & \multirow{2}{*}{0.62} & \multirow{2}{*}{0.13}\\
    & Normal & 624468 & 133815 & 133815 & & & & & & & \\

    \multirow{2}{*}{WADI}
    & Anomal. & 6984  & 1497  & 1496 
    & \multirow{2}{*}{123} &  \multirow{2}{*}{4.88} & \multirow{2}{*}{99.94} & \multirow{2}{*}{99.97} & \multirow{2}{*}{98.94} & \multirow{2}{*}{1.00} & \multirow{2}{*}{0.04} & \multirow{2}{*}{0.02}\\
    & Normal & 663152 & 142104 & 142105 & & & & & & & \\
    \bottomrule
    \end{tabular}
    \end{scriptsize}
    \vspace{-5pt}
    \caption{The column descriptors Train, Val., and Test indicate the number of samples in the training, validation, and test sets, respectively. Rejected and Approved descriptor specifies the number of rejected and approved loan applications in the LCLD dataset, whereas row descriptors Anomal. and Normal specifies the number of anomalous and normal samples. The Feat. column provides the size of the feature set, while the $\ell_2$-norm column reports the average $\ell_2$-norm of the examples in the respective dataset. The Accuracy and F1-score columns present the accuracy and F1-score of the target classifier used for the adversarial attack experiments. Finally, the columns TP, TN, FP, and FN represent the number of True Positives, True Negatives, False Positives, and False Negatives, respectively, reflecting the performance of the trained classifier employed in the adversarial attack.}
    \label{tab:dataset_n_clf}
\end{table*}

\begin{table*}[t!]
\centering
\begin{minipage}{0.60\textwidth}
\centering
\begin{footnotesize}
\begin{tabular}{lll}
\toprule
Layer & FFNN & 1D-CNN\\
\midrule
1 & Linear($d \rightarrow 1024$) 
  & Conv1D((*, d) $\rightarrow$ 512), kernel=5 \\
2 & Linear(1024 $\rightarrow$ 4096) 
  & AvgPool1D, kernel=3 \\
3 & Linear(4096 $\rightarrow$ 512) 
  & Conv1D((*, 512) $\rightarrow$ 1024), kernel=3 \\
4 & Linear(512 $\rightarrow$ 16) 
  & Flatten + Linear($\ast \rightarrow 2048$) \\
5 & Linear(16 $\rightarrow$ 1) 
  & Linear(2048 $\rightarrow$ 128) \\
6 & --- 
  & Linear(128 $\rightarrow$ 1) \\
Output & Raw output logits 
    & Raw output logits \\
\bottomrule
\end{tabular}
\end{footnotesize}
\vspace{-5pt}
\caption*{(a) Architecture of classifier with input dimension $d$}
\end{minipage}
\hfill
\begin{minipage}{0.38\textwidth}
\centering
\begin{footnotesize}
\begin{tabular}{ll}
\toprule
Hyperparameter & Value \\
\midrule
Optimizer & Adam \\
Learning Rate & 0.1 \\
Loss Function & BCEWithLogitsLoss \\
Mini Batch Size & $2^{10}$ \\
Epochs & 100 \\
Activation & ReLU \\
Nodes Connection & Fully Connected \\
\bottomrule
\end{tabular}
\end{footnotesize}
\vspace{-5pt}
\caption*{(b) Training hyperparameters}
\end{minipage}
\vspace{-5pt}
\caption{Table (a) provides the details about the architecture of the two neural nets used for the evaluation in the work. The final layer for both architectures returns the raw logits. The ReLU function is used as the activation function for each of the intermediate layers for both architectures. Note: * in the architecture of the 1D-CNN denotes dimensions that are variable or implicitly determined by the input dimension $d$ and the effects of preceding layers---e.g., sequence length after pooling or convolution. Table (b) provides the details about the hyperparameters used to train the classifiers. 
}
\label{tab:arch_and_hyperparams}
\end{table*}

\section{Dataset and Classifier}
\subsection{Dataset}
The selection of the datasets used for the evaluation is based on their relevance to the objective of the study, popularity, and ease of availability. The LCLD dataset was taken from the Kaggle website as one of the baseline MoEvA2 uses it's their work. We have downloaded the dataset from the Kaggle website just to compare the evaluation results with the other methods and have cited the download site in the main paper.

\paragraph{LCLD} The Lending Club Loan dataset is a benchmark dataset for evaluating credit risk modeling, loan default prediction, and financial decision-making algorithms in peer-to-peer (P2P) lending platforms. It was collected from the operations of Lending Club, a prominent online lending marketplace that connects borrowers and investors. The dataset contains detailed information on over 2 million loan applications issued between 2007 and 2018, including borrower demographics, loan attributes, credit history, and payment outcomes. Key features include FICO scores, debt-to-income ratios, employment length, loan purpose, and repayment status. Each loan record is labeled with the loan status (fully paid, charged off, or default), enabling supervised learning tasks such as binary and multi-class classification, as well as risk scoring and economic modeling. The dataset supports research in financial technology (FinTech), credit scoring, and responsible lending practices.

\paragraph{IDS} Intrusion Detection System dataset is a comprehensive benchmark dataset for evaluating intrusion detection methods in networked cyber-physical systems. It was generated by simulating realistic network traffic in a controlled testbed environment, incorporating benign activities as well as multiple types of contemporary cyberattacks. The dataset captures over 80 network traffic features, including flow-based and packet-based metrics, collected from various network devices and hosts over a 7-day period. Attack scenarios include DoS, DDoS, brute force, infiltration, botnets, and web attacks, among others, reflecting both volumetric and stealthy threats. The dataset contains labeled records indicating normal or attack traffic.

\paragraph{IoMT} Internet of Medical Things is a benchmark dataset for evaluating cybersecurity methods in the Internet of Medical Things (IoMT) environments. It was collected from a testbed comprising 40 IoMT devices (25 real and 15 simulated) operating over Wi-Fi, MQTT, and Bluetooth protocols. The dataset includes 18 distinct cyberattacks, categorized into five classes: DDoS, DoS, Recon, MQTT, and Spoofing. Data collection was performed using network taps to capture traffic between switches and IoMT devices, ensuring seamless and uninterrupted data acquisition. The dataset provides labeled records indicating normal or attack conditions, facilitating the development and evaluation of security solutions for IoMT systems.

\paragraph{SWaT} Secure Water Treatment dataset is a benchmark dataset for evaluating anomaly detection and intrusion detection methods in cyber-physical systems. It was collected from a scaled-down water treatment testbed that mimics the operations of a real industrial plant. The dataset includes readings from 51 sensors and actuators across 6 stages of the treatment process, captured under both normal operation and 36 simulated cyberattack scenarios. These attacks target various components such as pumps, valves, and chemical dosing systems, representing both insider and external threats. The dataset provides labeled measurements of physical process variables, making it suitable for research in industrial control system security.

\paragraph{WADI} Water Distribution dataset is a benchmark dataset for evaluating anomaly detection and intrusion detection methods in cyber-physical systems, in the context of water distribution networks. It was collected from a scaled-down water distribution testbed designed to replicate the behavior of a real-world urban water supply system. The dataset consists of 123 features, comprising 80 sensors and 43 actuators, recorded across various components such as pumps, valves, flow meters, and water tanks. The data spans two main phases: a normal operation period and an attack period featuring 15 days of simulated cyberattacks. These attacks were designed to compromise both the physical process and control logic, including sensor spoofing, actuator manipulation, and network-layer intrusions. Each record in the dataset is labeled as either normal or attack.

\begin{wrapfigure}{r}{0.4\textwidth}
\begin{minipage}{0.4\textwidth}
\centering
\begin{footnotesize}
\begin{tabular}{ll|ll}
\toprule
\textbf{Library} & \textbf{Version} & \textbf{Library} & \textbf{Version} \\
\midrule
torch           & 2.3.1            & numpy           & 1.26.4 \\
pymoo           & 0.4.2.2          & sympy           & 1.13.3 \\
scikit-learn    & 1.5.0            & pandas          & 2.2.2 \\
\bottomrule
\end{tabular}
\vspace{-10pt}
\caption{Essential Python libraries used in the implementation of the proposed and baseline methods.}
\label{tab:ml_dependencies}
\end{footnotesize}
\end{minipage}
\vspace{-20pt}
\end{wrapfigure}

\subsection{Software}
All the methods are implemented using Python programming (version: 3.12.3) environment over an Ubuntu server---details of the hardware are provided in the main article. The Table \ref{tab:ml_dependencies} highlights the core dependencies of the libraries used for the implementation of the proposed methods and baselines. The complete list of libraries is mentioned in the requirements.txt files---included with the artifacts.

\subsection{Classifier}
One-dimensional Convolutional Neural Networks (1D-CNN) and the Feed Forward Neural Networks are the two common architectures used over the structured dataset. The choice of layers---provided in Table \ref{tab:arch_and_hyperparams}(a) has given good classification results. However, over the LCLD, we tried both FFNN and 1D-CNN-based architecture, but the accuracy remains around 80\% for both choices. We choose to use 1D-CNN over the LCLD to diversify the choice of neural network architecture used for the evaluation in order to assess the effectiveness of the adversarial attacks method across a diverse choice of neural network architecture. Table \ref{tab:arch_and_hyperparams}(a) and \ref{tab:arch_and_hyperparams}(b) provide the details related to the architecture and the hyperparameters of the classifier used to train the network. Table \ref{tab:dataset_n_clf} includes the performance of the classifier over the respective dataset.

\begin{figure*}[t!]
\begin{subfigure}[t]{0.19\linewidth}
    \includegraphics[width=\linewidth]{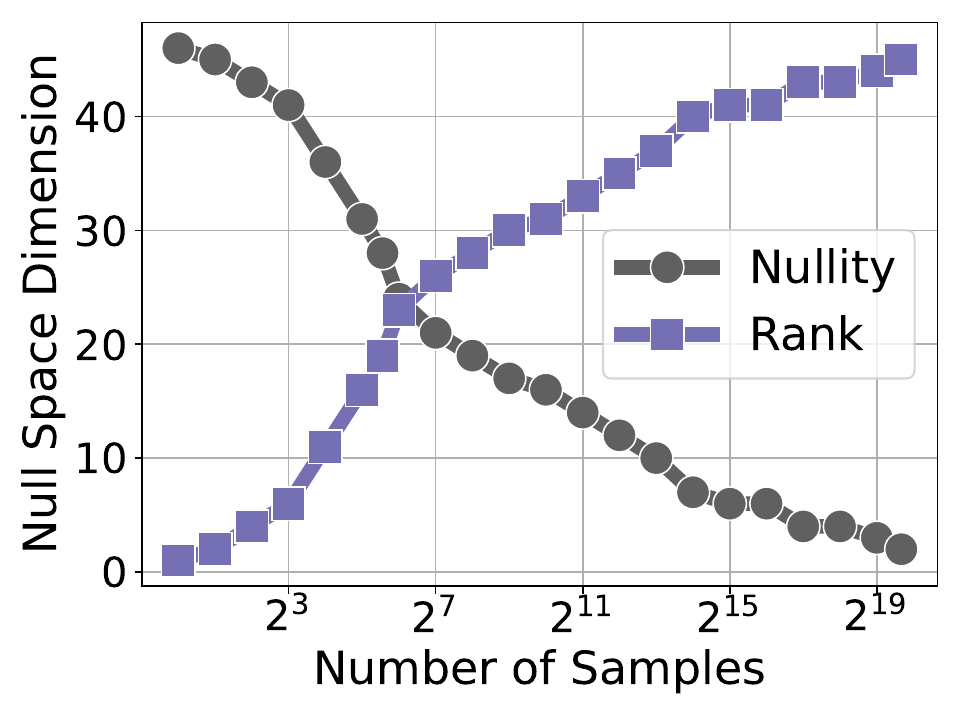}
    \vspace{-15pt}
    \caption{LCLD}
    \label{fig:LCLD_ns_dim}
\end{subfigure}
\hfill
\begin{subfigure}[t]{0.19\linewidth}
    \includegraphics[width=\linewidth]{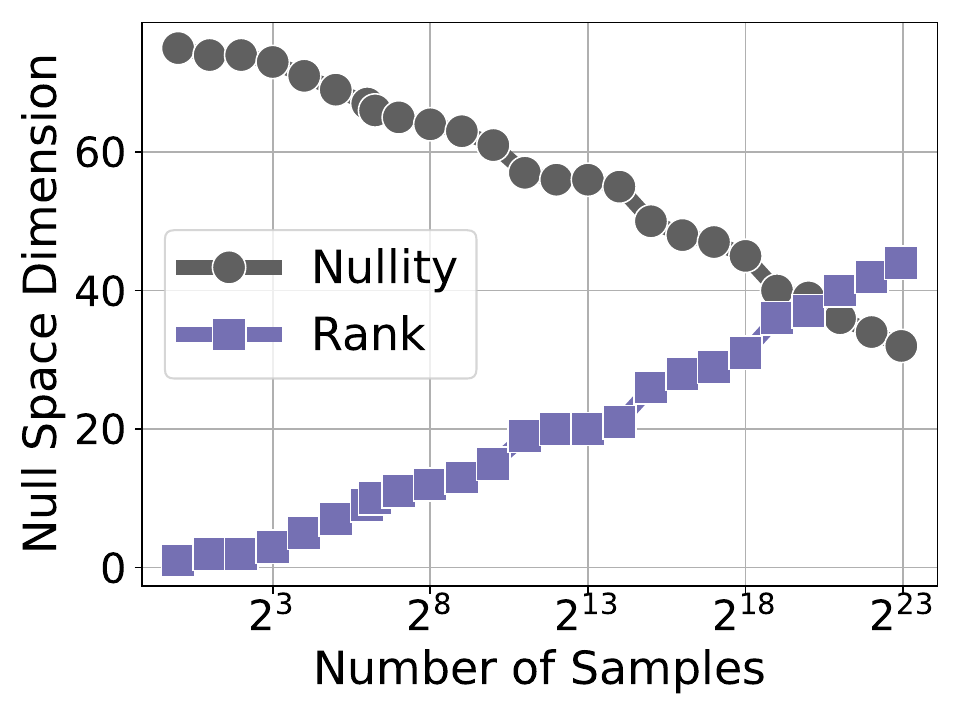}
    \vspace{-15pt}
    \caption{IDS}
    \label{fig:IDS_ns_dim}
\end{subfigure}
\hfill
\begin{subfigure}[t]{0.19\linewidth}
    \includegraphics[width=\linewidth]{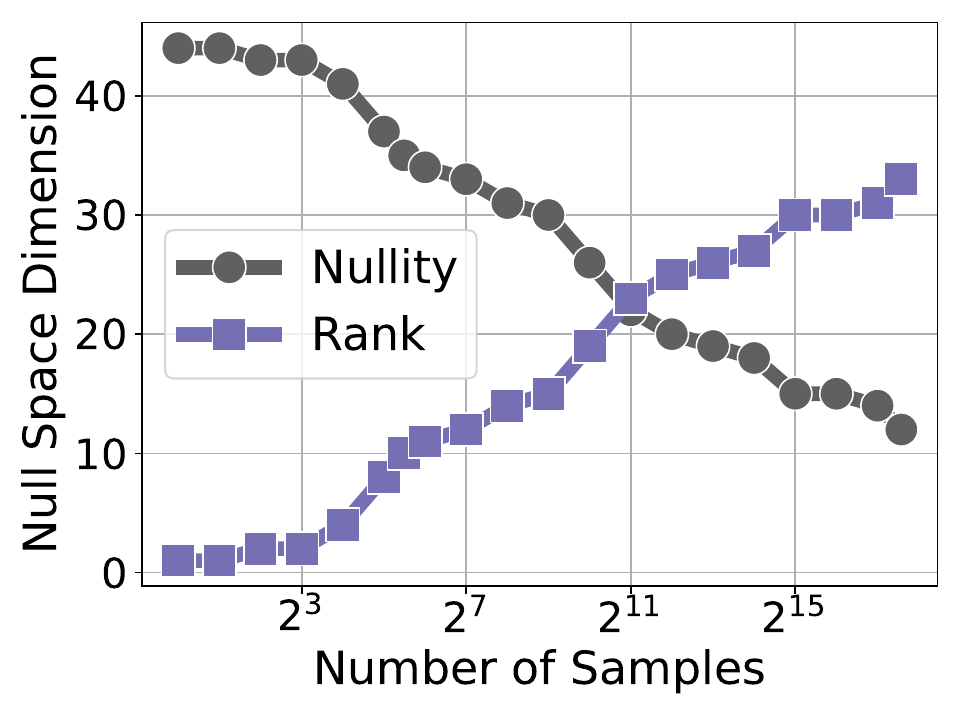}
    \vspace{-15pt}
    \caption{IoMT}
    \label{fig:IoMT_ns_dim}
\end{subfigure}
\hfill
\begin{subfigure}[t]{0.19\linewidth}
    \includegraphics[width=\linewidth]{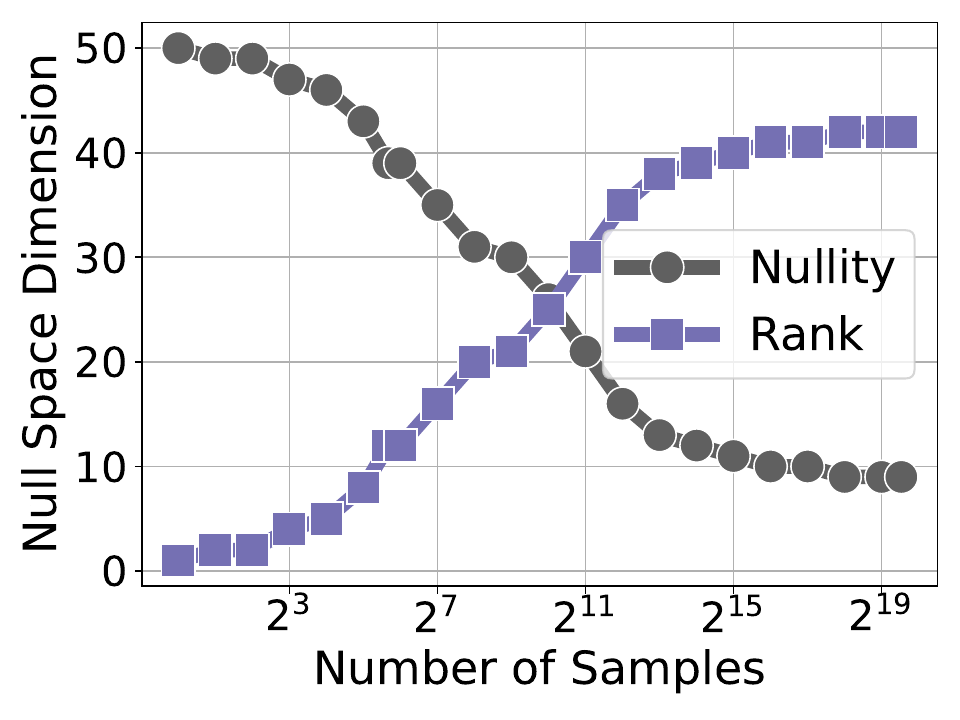}
    \vspace{-15pt}
    \caption{SWaT}
    \label{fig:SWaT_ns_dim}
\end{subfigure}
\hfill
\begin{subfigure}[t]{0.19\linewidth}
    \includegraphics[width=\linewidth]{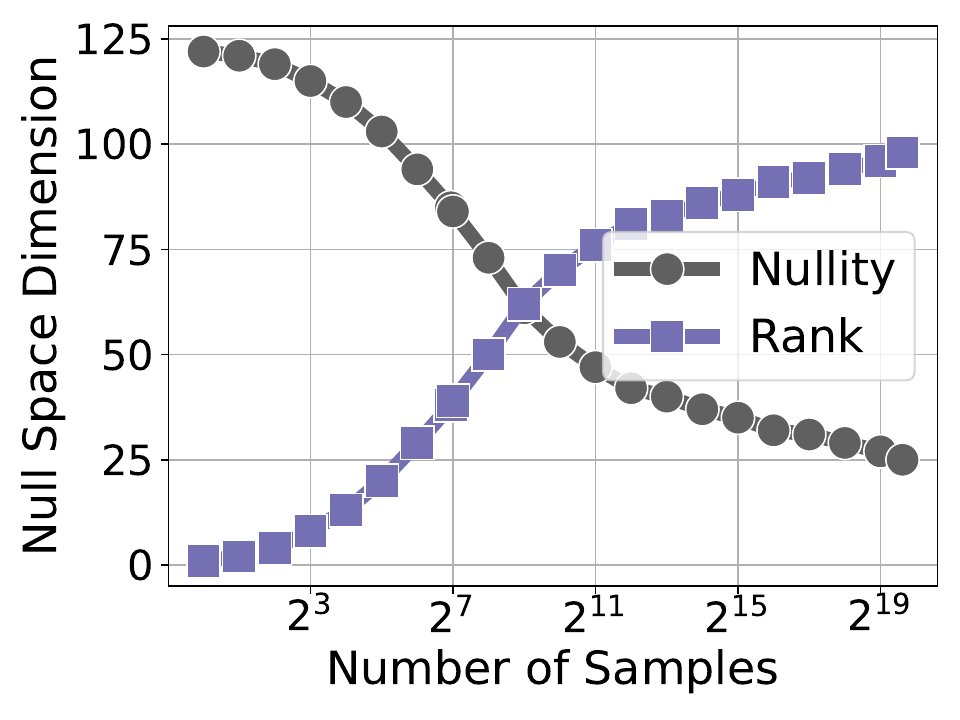}
    \vspace{-15pt}
    \caption{WADI}
\end{subfigure}
\vspace{-10pt}
\caption{The plot represents the rank of the column and null space. Normal samples are utilized to calculate the rank for the column and null space, as the normal samples encapsulate the normal behavior of the system. The plot reveals that even with an increasing number of inputs, the rank of the column space does not reach full rank, quantifying the amount of linear dependency present among the features.}
\label{fig:data_rank}
\end{figure*}

\begin{figure*}[t!]
\begin{center}
\begin{subfigure}[t]{0.19\linewidth}
    \includegraphics[width=\linewidth]{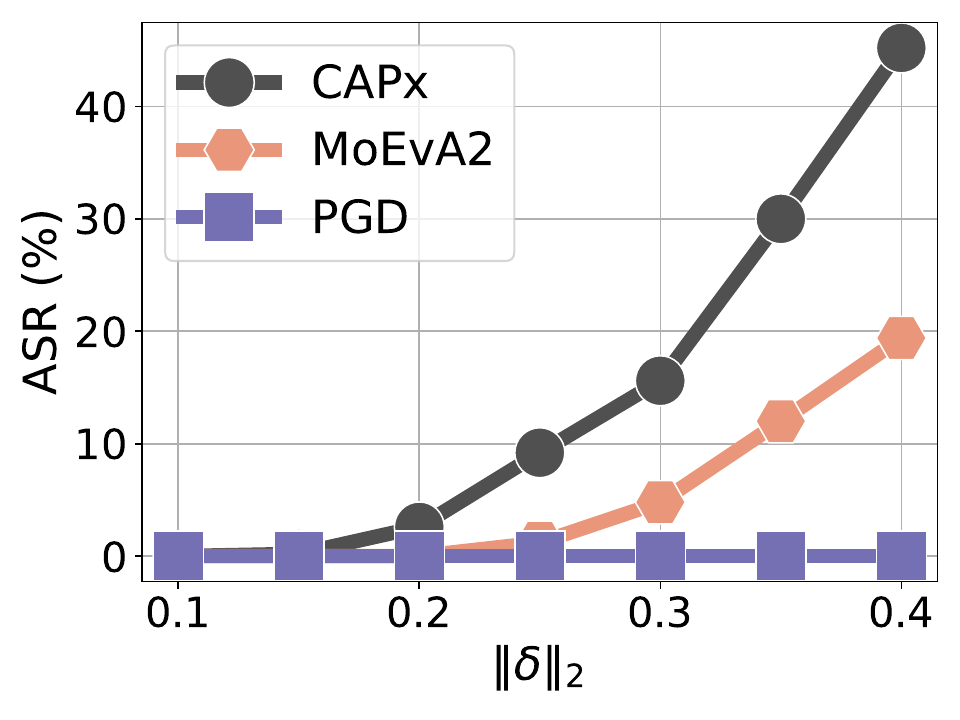}
    \vspace{-15pt}
    \caption{LCLD}
\end{subfigure}
\hfill
\begin{subfigure}[t]{0.19\linewidth}
    \includegraphics[width=\linewidth]{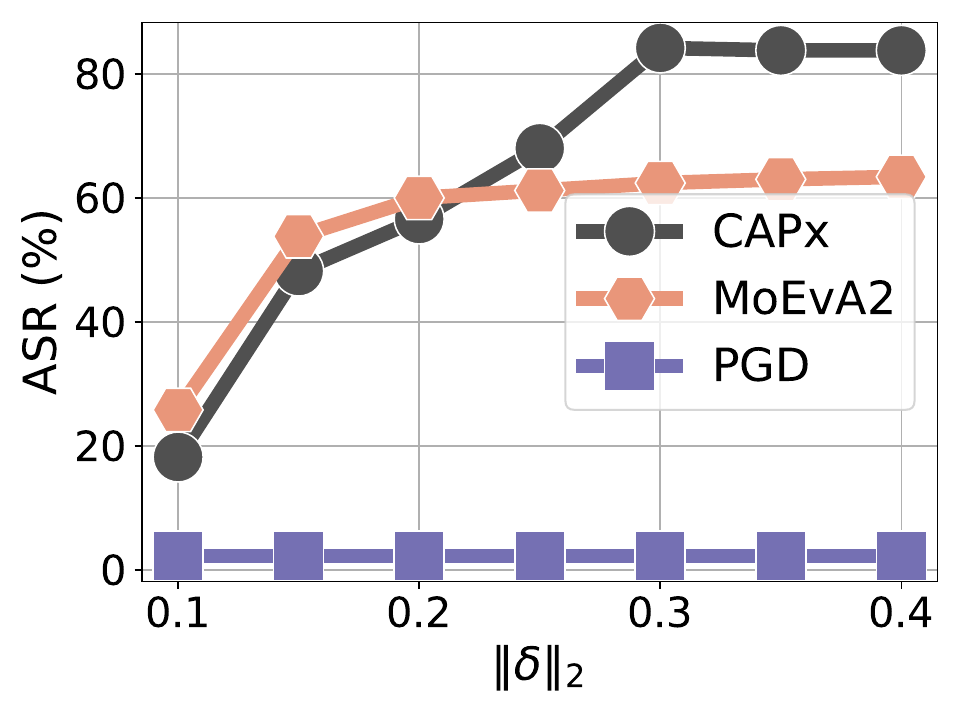}
    \vspace{-15pt}
    \caption{IDS}
\end{subfigure}
\hfill
\begin{subfigure}[t]{0.19\linewidth}
    \includegraphics[width=\linewidth]{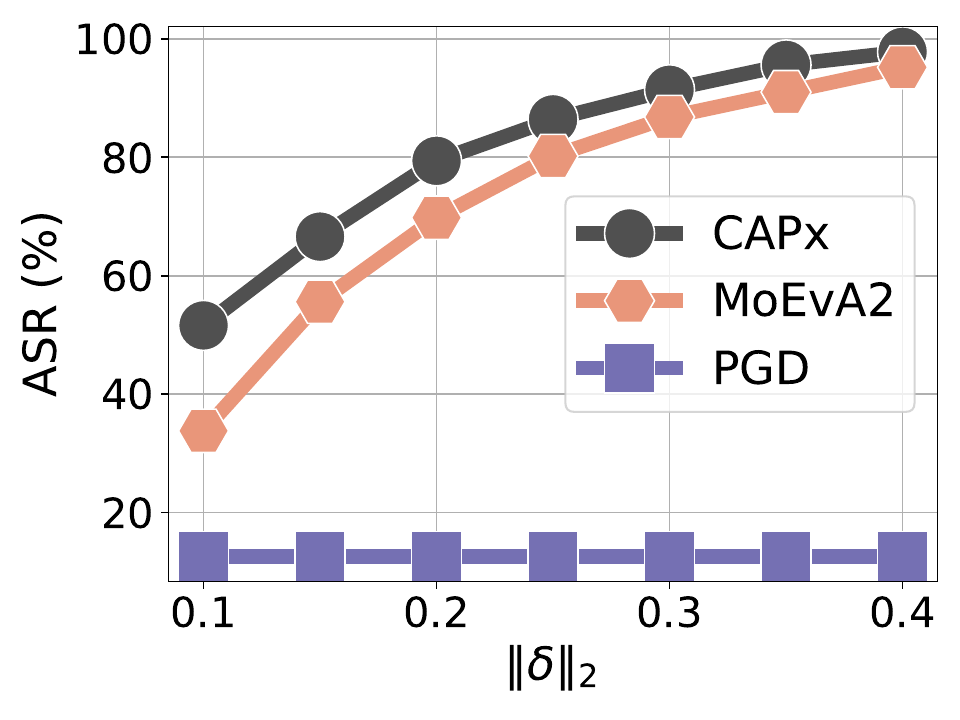}
    \vspace{-15pt}
    \caption{IoMT}
\end{subfigure}
\hfill
\begin{subfigure}[t]{0.19\linewidth}
    \includegraphics[width=\linewidth]{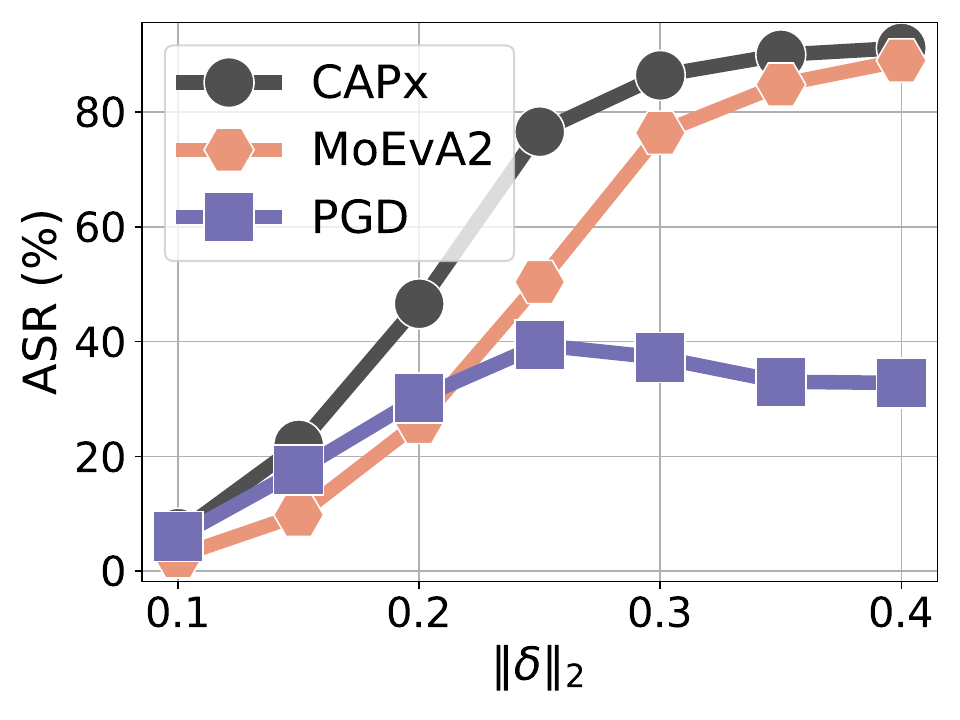}
    \vspace{-15pt}
    \caption{SWaT}
\end{subfigure}
\hfill
\begin{subfigure}[t]{0.19\linewidth}
    \includegraphics[width=\linewidth]{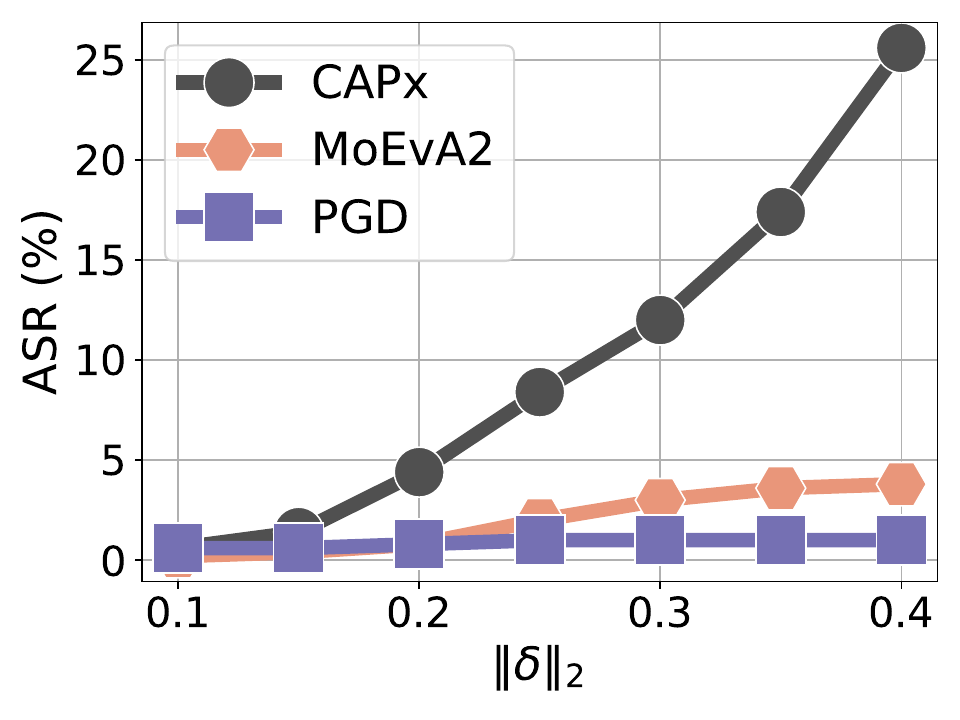}
    \vspace{-15pt}
    \caption{WADI}
\end{subfigure}
\end{center}
\vspace{-20pt}
\caption{
    Evaluation results to observe the adversarial efficacy of the \textit{individual perturbation} methods under the norm budget.
}
\label{fig:x_asr_vs_norm}
\end{figure*}

\section{Experimental Results}
One of the hyperparameters of the attack method is the learning rate or step size ($\alpha$), which is set using cross-validation for the evaluation. For the LCLD and IDS dataset, we have found that the Adam-based update with $\alpha=0.01$ for methods \capx and \capX gives the better performance, whereas over the rest of the other datasets, we have found that a fixed learning rate $\alpha=0.01$ with penalty fixed to $1$ works well. 
DeepFool and MoEvA2 do not require any explicit learning rate. For the PGD and ConAML, the learning rate is fixed to $0.01$---set based on the cross-validation. 

Figure \ref{fig:x_asr_vs_norm} observes the adversarial efficacy of the individual perturbation methods over the allowed $\ell_2$-norm budget with maximum allowed iteration fixed to 300 for each method---hyperparameter specific to MoEvA2: population size set to 64 and offspring size set to 32. 

\begin{figure*}[t!]
\begin{center}
\begin{subfigure}[t]{0.19\linewidth}
    \includegraphics[width=\linewidth]{Plots/X_Plot/Mod_X_VS_Time/LCLD_Mod_X_vs_Time.pdf}
    \vspace{-15pt}
    \caption{LCLD}
    \label{fig:LCLD_mclf_time}
\end{subfigure}
\hfill
\begin{subfigure}[t]{0.19\linewidth}
    \includegraphics[width=\linewidth]{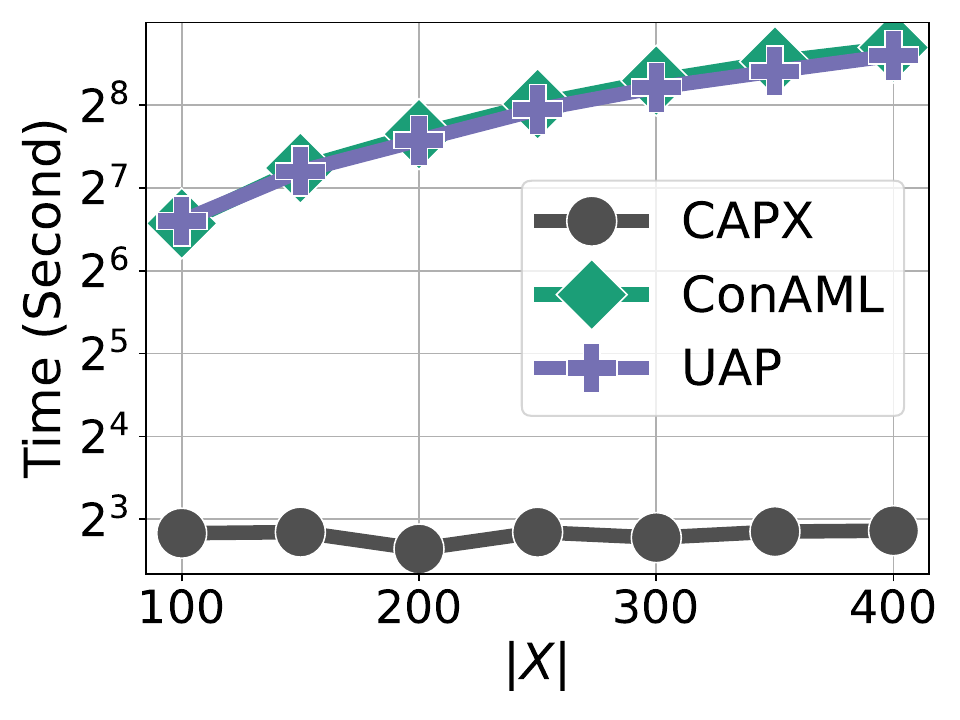}
    \vspace{-15pt}
    \caption{IDS}
    \label{fig:IDS_mclf_time}
\end{subfigure}
\hfill
\begin{subfigure}[t]{0.19\linewidth}
    \includegraphics[width=\linewidth]{Plots/X_Plot/Mod_X_VS_Time/IoMT_Mod_X_vs_Time.pdf}
    \vspace{-15pt}
    \caption{IoMT}
    \label{fig:IoMT_mclf_time}
\end{subfigure}
\hfill
\begin{subfigure}[t]{0.19\linewidth}
    \includegraphics[width=\linewidth]{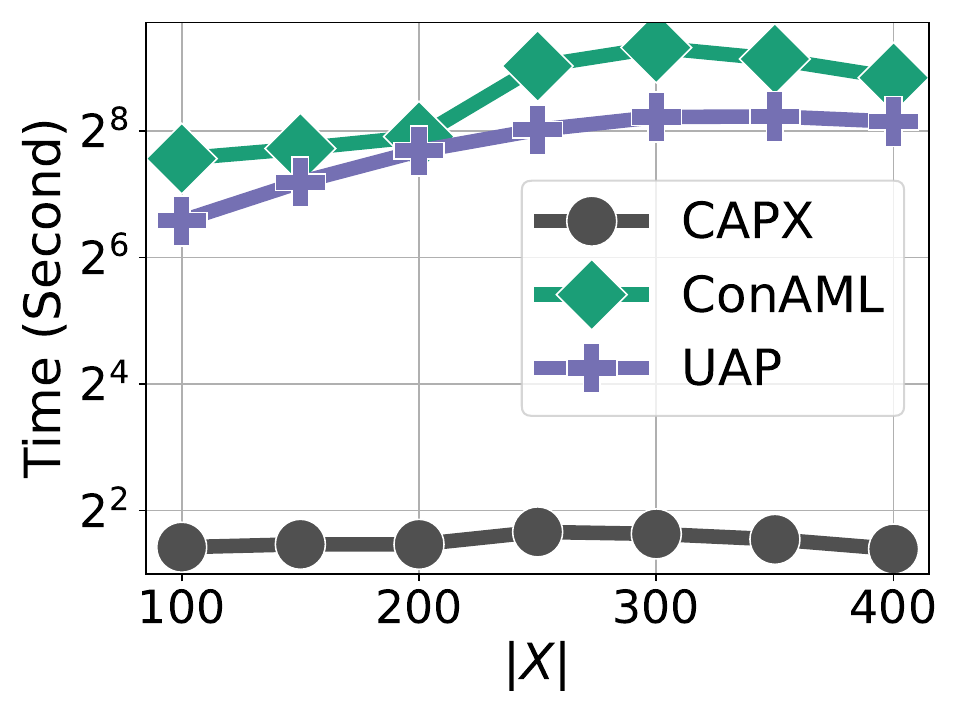}
    \vspace{-15pt}
    \caption{SWaT}
    \label{fig:SWaT_mclf_time}
\end{subfigure}
\hfill
\begin{subfigure}[t]{0.19\linewidth}
    \includegraphics[width=\linewidth]{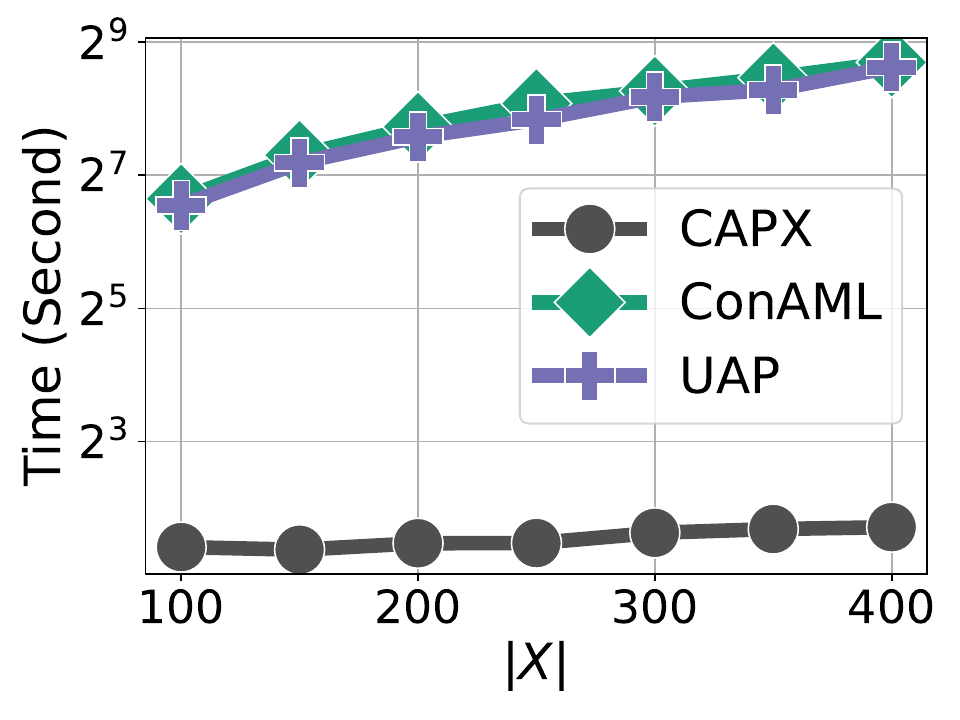}
    \vspace{-15pt}
    \caption{WADI}
    \label{fig:wadi_mclf_time}
\end{subfigure}
\end{center}
\vspace{-20pt}
\caption{
    Note: The y-axis uses a logarithmic scale with base 2. \capX is highly time-efficient in comparison to baselines.
}
\label{fig:asr_vs_time}
\end{figure*}

\begin{figure*}[t!]
    \centering
    \begin{subfigure}{0.19\textwidth}
        \includegraphics[width=\linewidth]{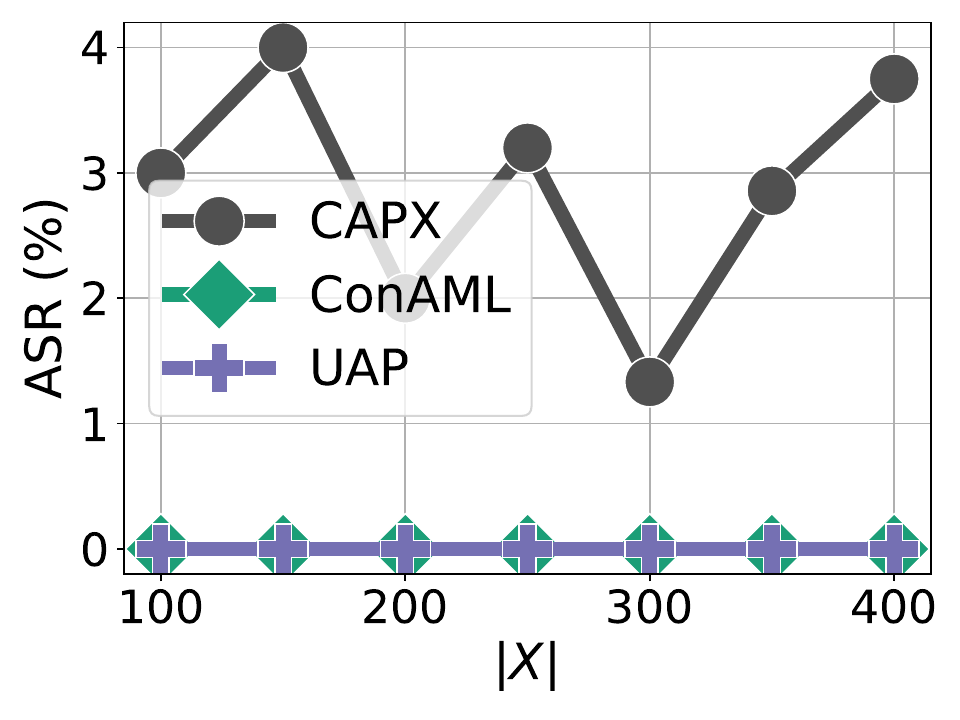}
    \end{subfigure}
    \begin{subfigure}{0.19\textwidth}
        \includegraphics[width=\linewidth]{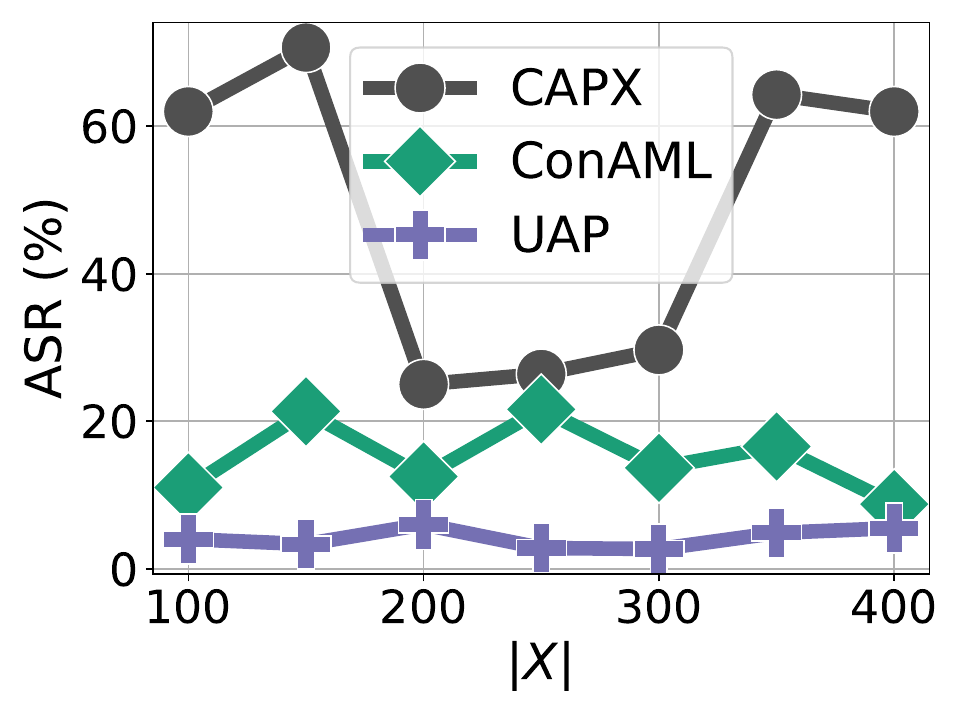}
    \end{subfigure}
    \begin{subfigure}{0.19\textwidth}
        \includegraphics[width=\linewidth]{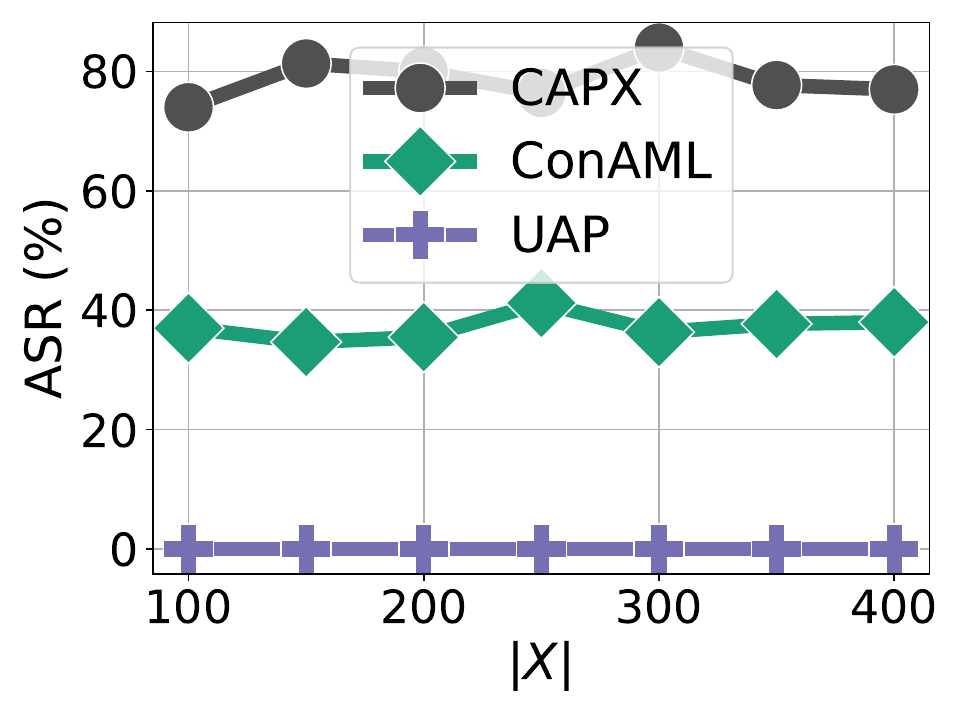}
    \end{subfigure}
    \begin{subfigure}{0.19\textwidth}
        \includegraphics[width=\linewidth]{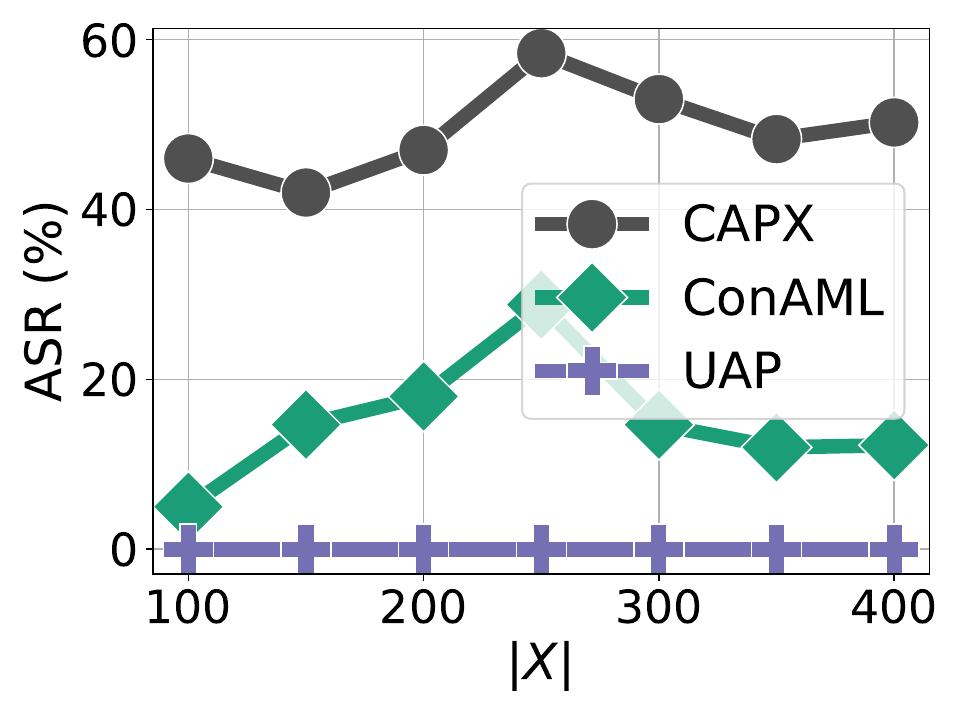}
    \end{subfigure}
    \begin{subfigure}{0.19\textwidth}
        \includegraphics[width=\linewidth]{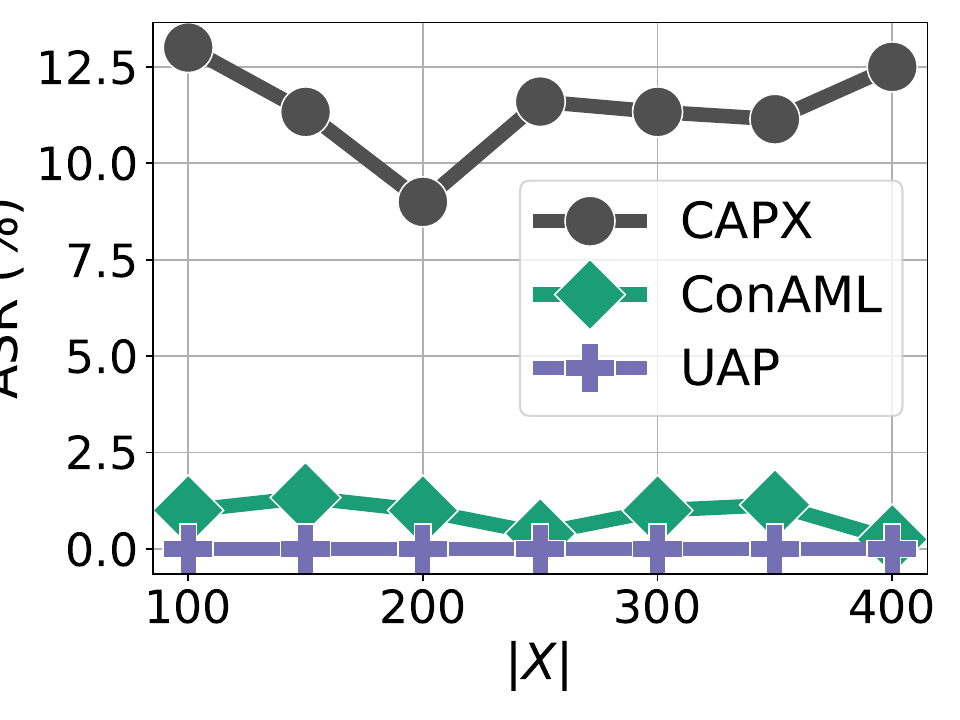}
    \end{subfigure}

    \begin{subfigure}{0.19\textwidth}
        \includegraphics[width=\linewidth]{Plots/X_Plot/Mod_X_VS_ASR_T/LCLD_Mod_X_vs_ASR_T.pdf}
        \vspace{-15pt}
        \caption{LCLD}
    \end{subfigure}
    \begin{subfigure}{0.19\textwidth}
        \includegraphics[width=\linewidth]{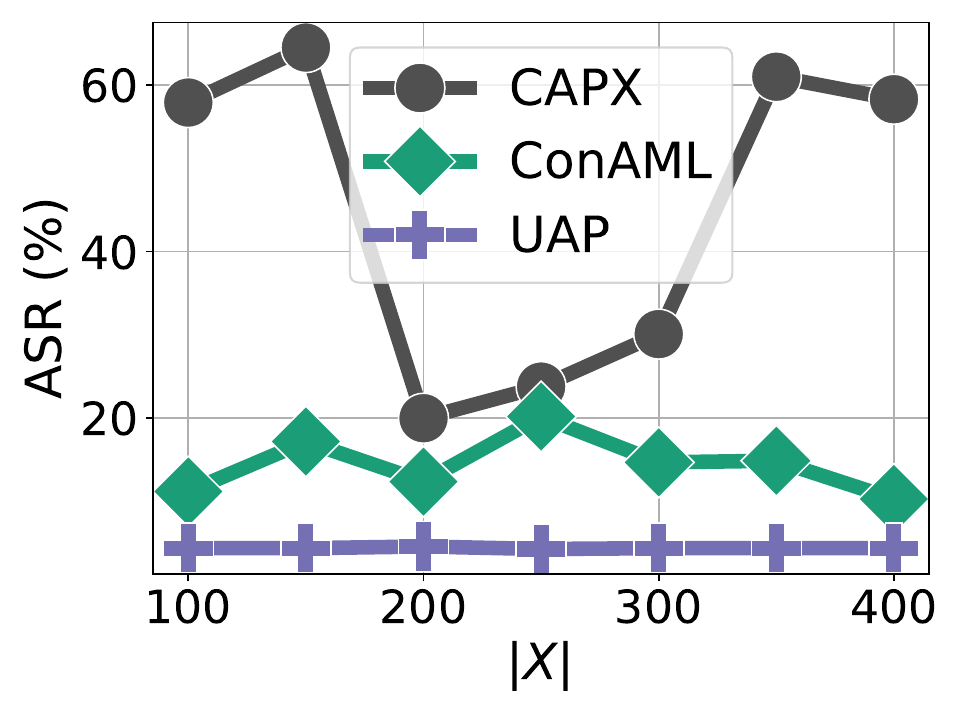}
        \caption{IDS}
    \end{subfigure}
    \begin{subfigure}{0.19\textwidth}
        \includegraphics[width=\linewidth]{Plots/X_Plot/Mod_X_VS_ASR_T/IoMT_Mod_X_vs_ASR_T.pdf}
        \vspace{-15pt}
        \caption{IoMT}
    \end{subfigure}
    \begin{subfigure}{0.19\textwidth}
        \includegraphics[width=\linewidth]{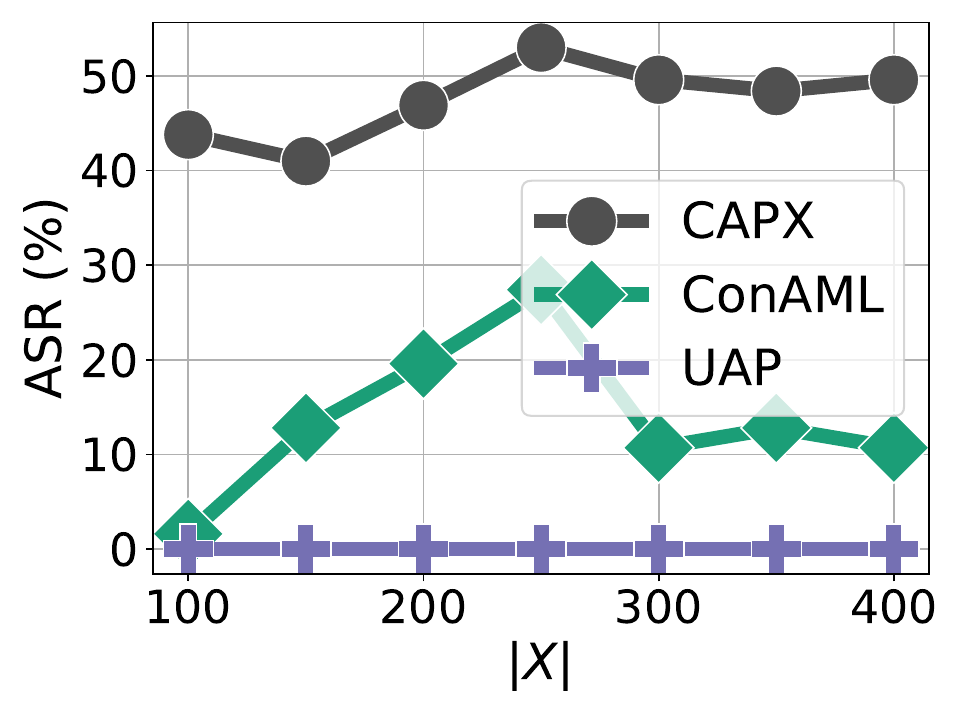}
        \vspace{-15pt}
        \caption{SWaT}
    \end{subfigure}
    \begin{subfigure}{0.19\textwidth}
        \includegraphics[width=\linewidth]{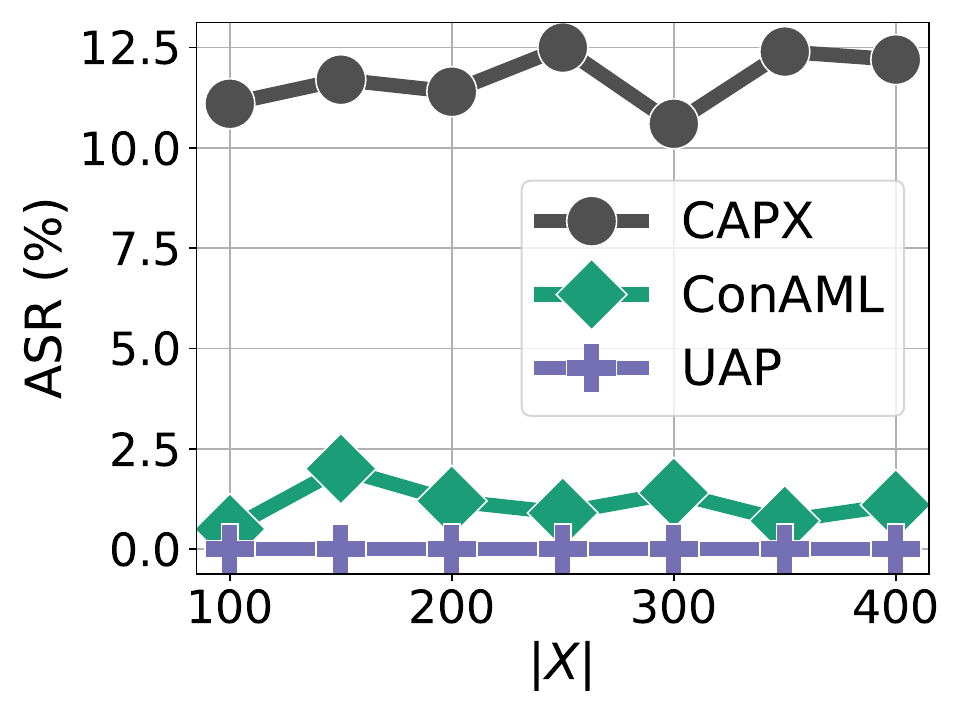}
        \vspace{-15pt}
        \caption{WADI}
    \end{subfigure}
    \vspace{-10pt}
    \caption{
        Experiment to observe the dependence of the ASR of the \textit{universal perturbation} methods over the samples in the training set $\bm{X}$. Subplots over the top row represent the ASRs of the methods over the training set $\bm{X}$, whereas the subplots over the bottom row observe the generalization over the unseen test examples.
    }
    \label{fig:app_X_asr_vs_X}
\end{figure*}

\begin{figure*}[tb]
    \centering
    \begin{subfigure}{0.19\textwidth}
        \includegraphics[width=\linewidth]{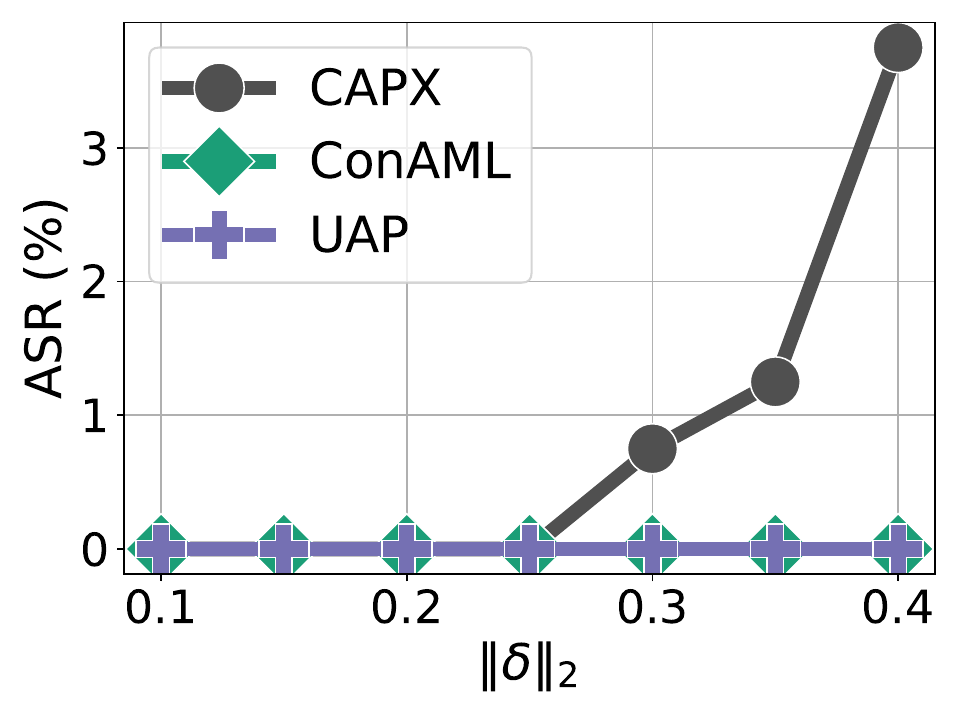}
    \end{subfigure}
    \begin{subfigure}{0.19\textwidth}
        \includegraphics[width=\linewidth]{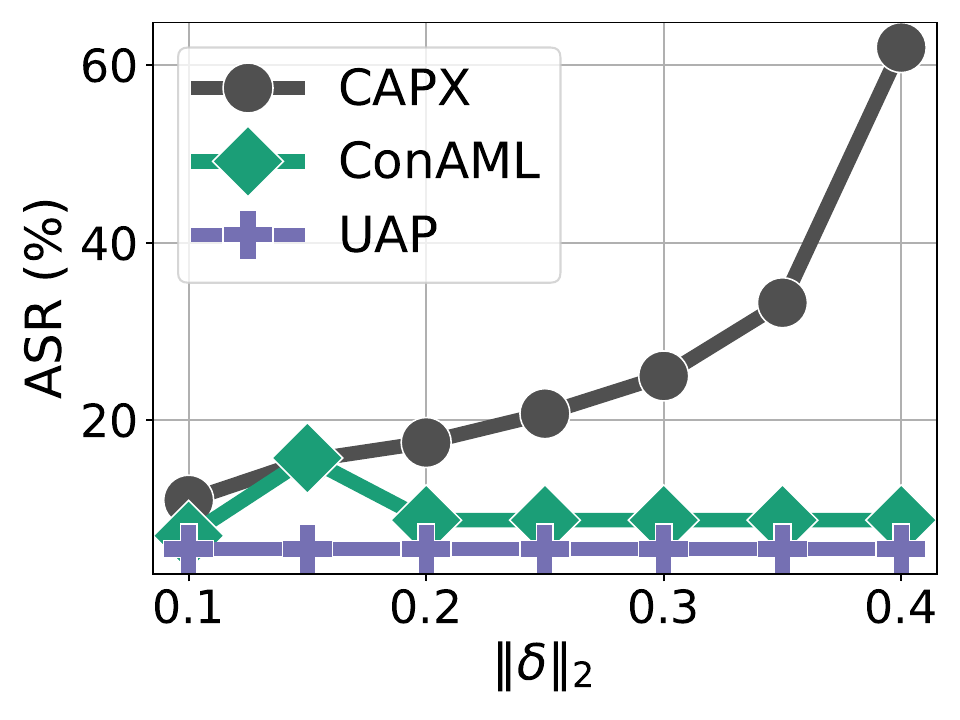}
    \end{subfigure}
    \begin{subfigure}{0.19\textwidth}
        \includegraphics[width=\linewidth]{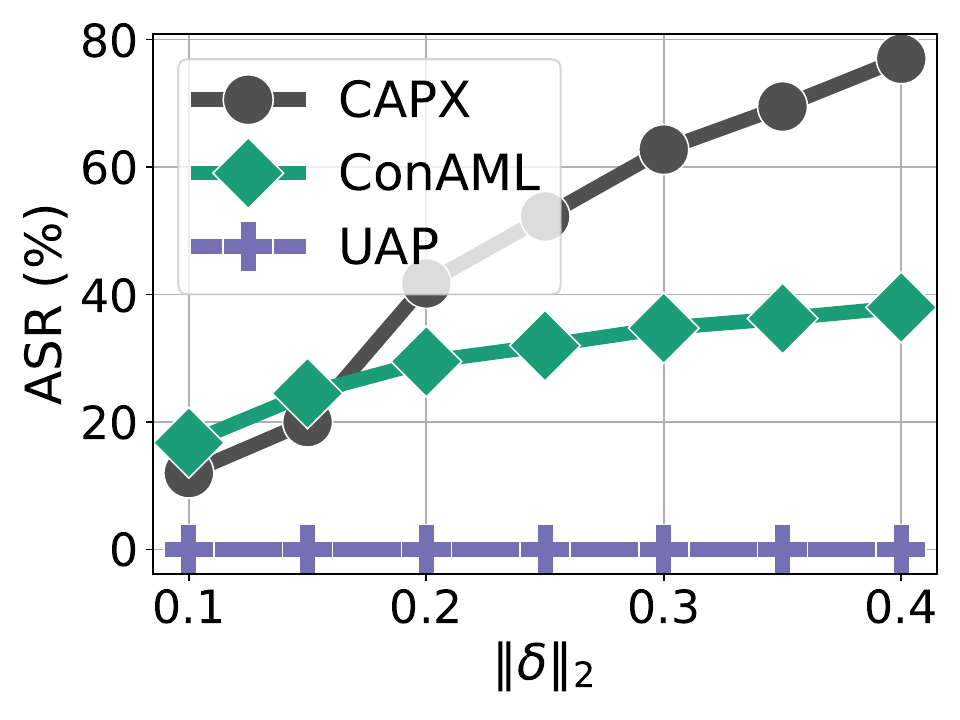}
    \end{subfigure}
    \begin{subfigure}{0.19\textwidth}
        \includegraphics[width=\linewidth]{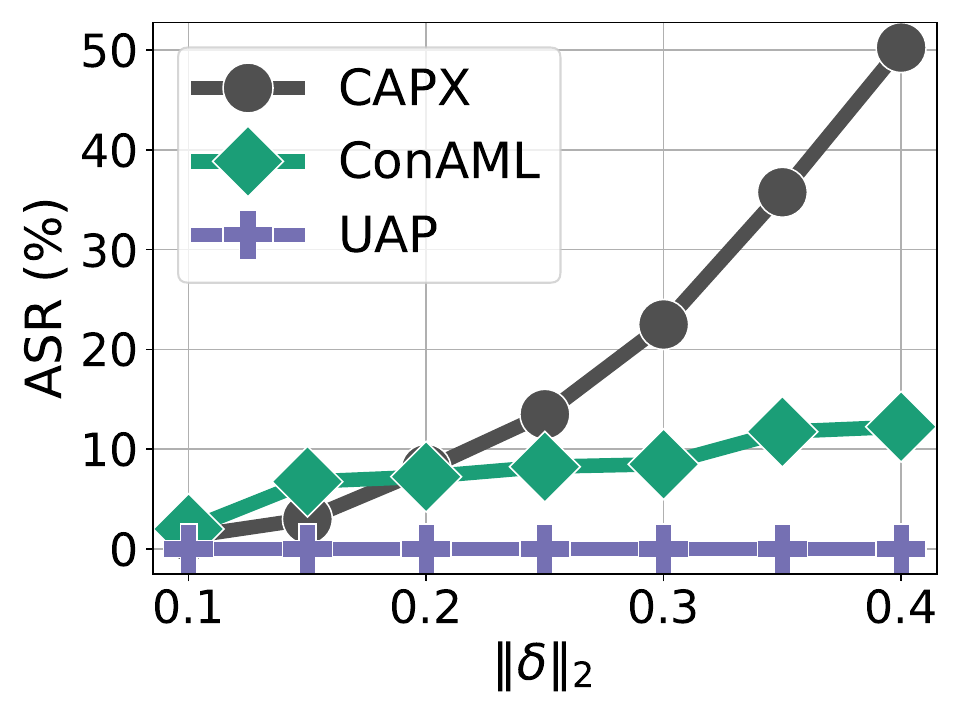}
    \end{subfigure}
    \begin{subfigure}{0.19\textwidth}
        \includegraphics[width=\linewidth]{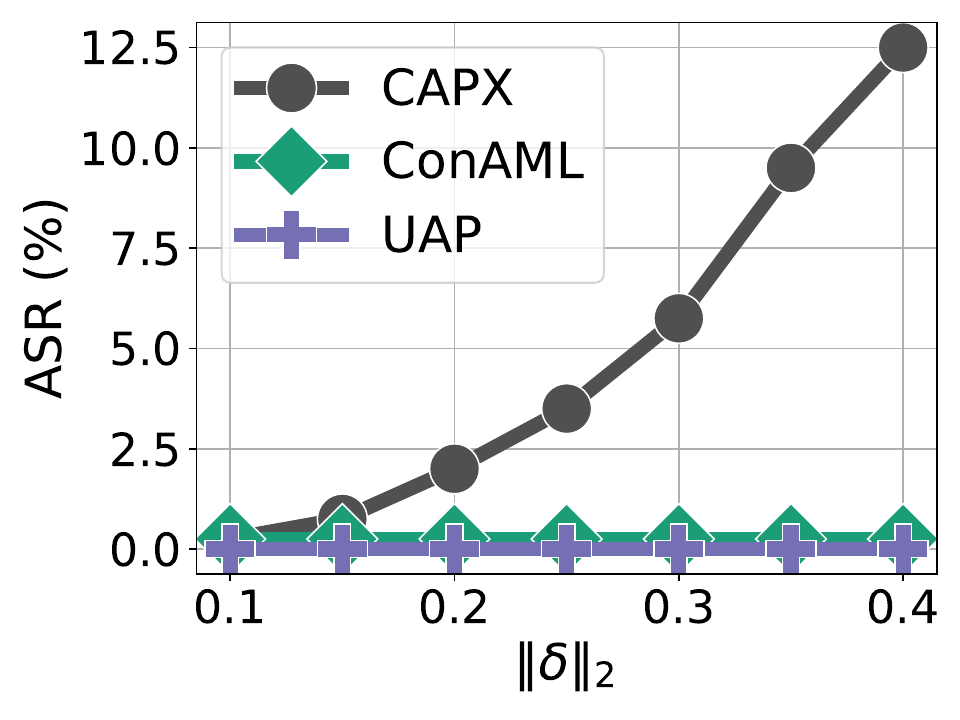}
    \end{subfigure}

    \begin{subfigure}{0.19\textwidth}
        \includegraphics[width=\linewidth]{Plots/X_Plot/Norm_VS_ASR_T/LCLD_Norm_vs_ASR_T.pdf}
        \vspace{-15pt}
        \caption{LCLD}
    \end{subfigure}
    \begin{subfigure}{0.19\textwidth}
        \includegraphics[width=\linewidth]{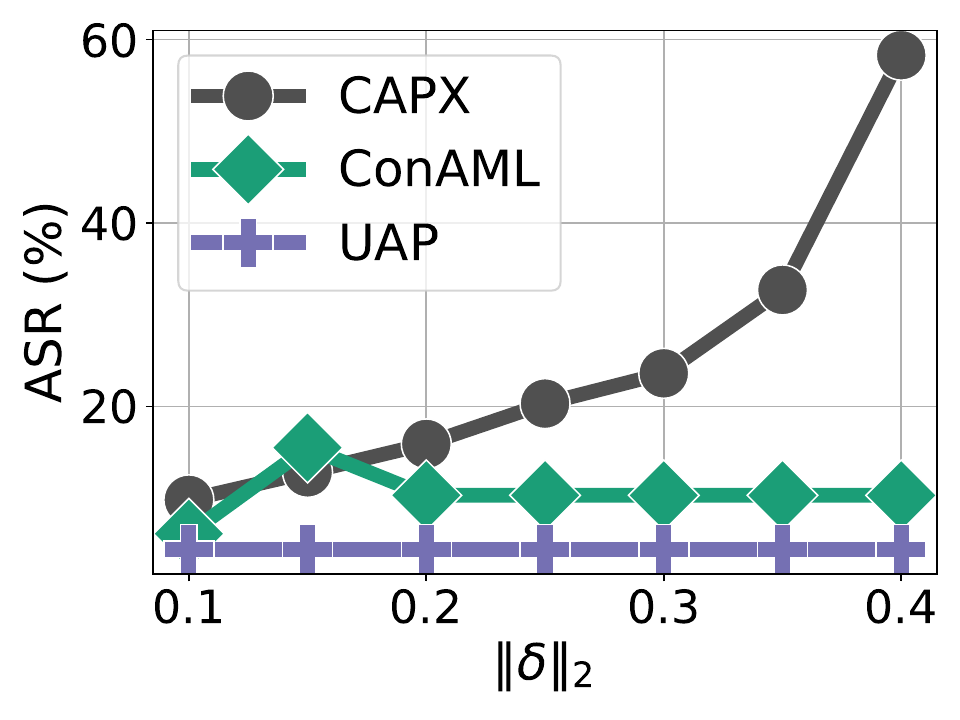}
        \vspace{-15pt}
        \caption{IDS}
    \end{subfigure}
    \begin{subfigure}{0.19\textwidth}
        \includegraphics[width=\linewidth]{Plots/X_Plot/Norm_VS_ASR_T/IoMT_Norm_vs_ASR_T.pdf}
        \vspace{-15pt}
        \caption{IoMT}
    \end{subfigure}
    \begin{subfigure}{0.19\textwidth}
        \includegraphics[width=\linewidth]{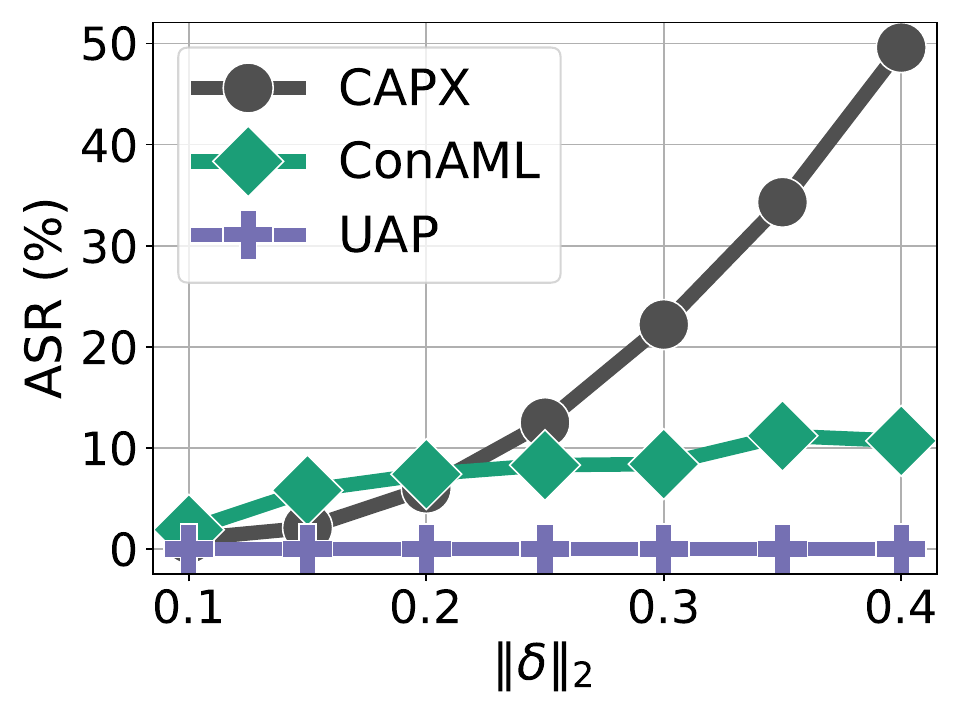}
        \vspace{-15pt}
        \caption{SWaT}
    \end{subfigure}
    \begin{subfigure}{0.19\textwidth}
        \includegraphics[width=\linewidth]{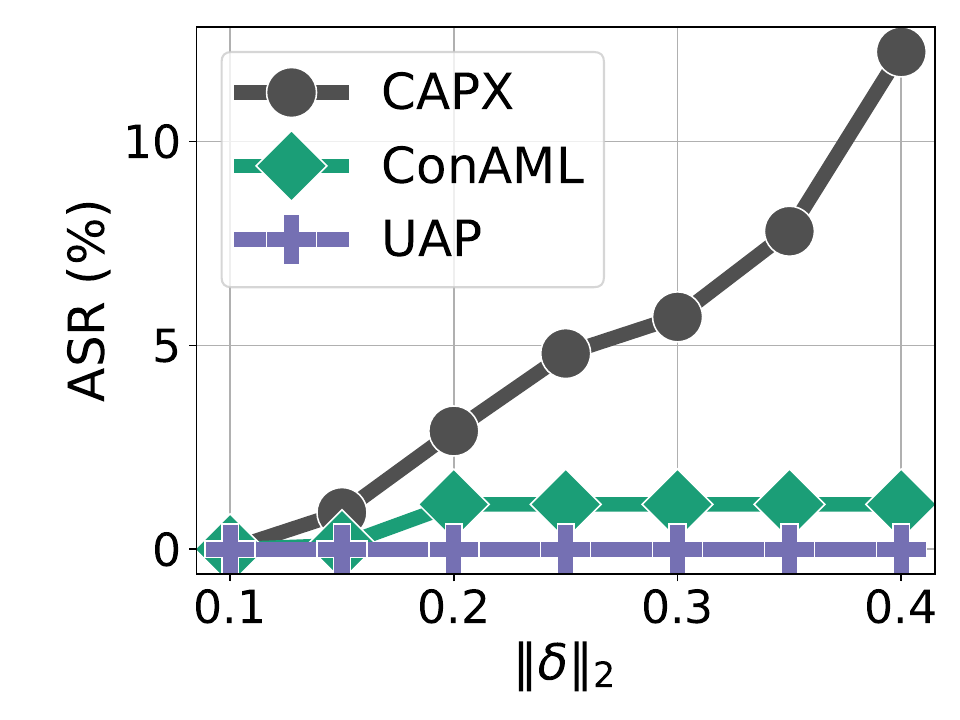}
        \vspace{-15pt}
        \caption{WADI}
    \end{subfigure}
    \vspace{-10pt}
    \caption{
        Experimental results to observe the adversarial efficacy of the \textit{universal perturbation} methods over the norm budget of the perturbation. The subplots over the top rows represent the ASR over the training set $\bm{X}$, whereas the subplot over the bottom row represents the ASR over the test examples.
    }
    \label{fig:app_X_asr_vs_l2}
\end{figure*}

Figure \ref{fig:asr_vs_time} observes the time taken by each of the universal methods as a function of the size of the training set $\bm{X}$.
Figure \ref{fig:app_X_asr_vs_X} and \ref{fig:app_X_asr_vs_l2} observe the ASR of the universal perturbation as a function of $\bm{X}$ and norm budget over the perturbation, respectively. Performance of both \capx and \capX is superior in comparison to baselines.

\end{document}